\documentclass{article}

\usepackage{microtype}
\usepackage{graphicx}
\usepackage{subfigure}
\usepackage{booktabs} % for professional tables
\usepackage{enumitem}

\usepackage{hyperref}

\usepackage[accepted]{icml2023}

\usepackage{svg}
\usepackage{tikz}
\usepackage{amsmath}
\usepackage{amssymb}
\usepackage{algorithm}
\usepackage[algo2e]{algorithm2e} 
\usepackage{algorithmic}
\usepackage{mathtools}
\usepackage{amsthm}
\usepackage{pythonhighlight}
\usepackage[toc,page]{appendix}

\theoremstyle{plain}
\newtheorem{theorem}{Theorem}[section]
\newtheorem{proposition}[theorem]{Proposition}
\newtheorem{lemma}[theorem]{Lemma}
\newtheorem{corollary}[theorem]{Corollary}
\theoremstyle{definition}

\theoremstyle{remark}

\usepackage[utf8]{inputenc} % allow utf-8 input
\usepackage[T1]{fontenc}    % use 8-bit T1 fonts
\usepackage{hyperref}       % hyperlinks
\usepackage{url}            % simple URL typesetting
\usepackage{booktabs}       % professional-quality tables
\usepackage{amsfonts}       % blackboard math symbols
\usepackage{nicefrac}       % compact symbols for 1/2, etc.
\usepackage{microtype}      % microtypography
\usepackage{xcolor}         % colors
\usepackage{multirow}
\usepackage{wrapfig}
\usepackage{graphicx}

\usepackage{wrapfig}
\usepackage{booktabs}
\usepackage{amsmath,amssymb,amsfonts}
\usepackage{amsthm}
\usepackage{graphicx}
\usepackage{multirow}
\usepackage[para]{footmisc}

\newcommand{\mcG}{\mathcal{G}} % Graph
\newcommand{\mcV}{\mathcal{V}} % Set of nodes 

\newcommand{\mcE}{\mathcal{E}} % Set of edges
\newcommand{\mcR}{\mathcal{R}}
\newcommand{\mcN}{\mathcal{N}}

\newcommand{\mbY}{\mathbf{Y}} % Unfolding vector
\newcommand{\mbX}{\mathbf{X}} % Input matrix
\newcommand{\mbD}{\mathbf{D}} % Degree matrix
\newcommand{\mbI}{\mathbf{I}} % Identity matrix
\newcommand{\mbZ}{\mathbf{Z}} % Normalizing constant matrix

\newcommand{\bprox}{\textbf{prox}} % proximal operator

\usepackage[capitalize,noabbrev]{cleveref}

\input{def.set}

\usepackage[textsize=tiny]{todonotes}

\icmltitlerunning{From Hypergraph Energy Functions to Hypergraph Neural Networks}

\begin{document}

\twocolumn[
\icmltitle{From Hypergraph Energy Functions to Hypergraph Neural Networks}




\icmlsetsymbol{wkdaamz}{\dag}

\begin{icmlauthorlist}
\icmlauthor{Yuxin Wang}{fudan,ling,wkdaamz}
\icmlauthor{Quan Gan}{aws}
\icmlauthor{Xipeng Qiu}{fudan,pengcheng}
\icmlauthor{Xuanjing Huang}{fudan,scicivc}
\icmlauthor{David Wipf}{aws}
\end{icmlauthorlist}

\icmlaffiliation{fudan}{School of Computer Science, Fudan University}
\icmlaffiliation{aws}{Amazon}

\icmlaffiliation{pengcheng}{Peng Cheng Laboratory}
\icmlaffiliation{ling}{Institute of Modern Languages and Linguistics, Fudan University}
\icmlaffiliation{scicivc}{Shanghai Collaborative Innovation Center of Intelligent Visual Computing}
% \icmlaffiliation{aws}{AWS AI, Shanghai, China.}

\icmlcorrespondingauthor{Yuxin Wang}{wangyuxin21@m.fudan.edu.cn}
\icmlcorrespondingauthor{Quan Gan}{quagan@amazon.com}
\icmlcorrespondingauthor{Xipeng Qiu}{xpqiu@fudan.edu.cn}
\icmlcorrespondingauthor{Xuanjing Huang}{xjhuang@fudan.edu.cn}
\icmlcorrespondingauthor{David Wipf}{davidwipf@gmail.com}
\icmlkeywords{Machine Learning, ICML}

\vskip 0.3in
]



\printAffiliationsAndNotice{\textsuperscript{\dag}Work completed during an internship at the AWS Shanghai AI Lab.}

\begin{abstract}
Hypergraphs are a powerful abstraction for representing higher-order interactions between entities of interest.  To exploit these relationships in making downstream predictions, a variety of hypergraph neural network architectures have recently been proposed, in large part building upon precursors from the more traditional graph neural network (GNN) literature.  Somewhat differently, in this paper we begin by presenting an expressive family of parameterized, hypergraph-regularized energy functions.  We then demonstrate how minimizers of these energies effectively serve as node embeddings that, when paired with a parameterized classifier, can be trained end-to-end via a supervised bilevel optimization process.  Later, we draw parallels between the implicit architecture of the predictive models emerging from the proposed bilevel hypergraph optimization, and existing GNN architectures in common use.  Empirically, we demonstrate state-of-the-art results on various hypergraph node classification benchmarks. Code is available at \href{https://github.com/yxzwang/PhenomNN}{https://github.com/yxzwang/PhenomNN}.

\end{abstract}

\section{Introduction}
Hypergraphs represent a natural extension of graphs, whereby each hyperedge can link an arbitrary number of hypernodes (or nodes for short).  This flexibility more directly facilitates the modeling of higher-order relationships between entities \cite{chien2022you,benson2016higher,benson2017spacey} leading to strong performance in diverse real-world situations \cite{agarwal2005beyond,li2017inhomogeneous,hgnn,UniGNN}.  Currently, hypergraph-graph-based modeling techniques frequently rely, either implicitly or explicitly, on some type of expansion (e.g., clique, star), which effectively converts the hypergraph into a regular graph with a new edge  set and possibly additional nodes as well.  For example, one approach is to first extract a particular expansion graph and then build a graph neural network (GNN) model on top of it \cite{Zhang2022HypergraphCN}.  

We instead adopt a different starting point that both allows us to incorporate multiple expansions if needed, but also transparently explore the integrated role of each expansion within a unified framework.  To accomplish this, our high-level strategy is to first define a family of parameterized hypergraph energy functions, with regularization factors that we later show closely align with popular existing expansions.  We then demonstrate how the minimizers of such energy functions can be treated as learnable node embeddings and trained end-to-end via a bilevel optimization process.  Namely, the lower-level minimization process produces optimal features contigent on a given set of parameters, while the higher-level process trains these parameters (and hence the features they influence) w.r.t.~downstream node classification tasks.

To actualize this goal, after presenting related work in Section \ref{sec:related_work}, we provide relevant background and notation w.r.t.~hypergraphs in Section \ref{sec:notation}.  The remainder of the paper then presents our primary contributions, which can be summarized as follows:

\begin{itemize}
\item We present a general class of hypergraph-regularized energy functions in Section \ref{sec:family} and elucidate their relationship with traditional hypergraph expansions that have been previously derived from spectral graph theory.

\item We demonstrate how minimizers of these energy functions can serve as principled, trainable features for hypergraph prediction tasks in Sections \ref{sec:optimization} and \ref{sec:connections}.  And by approximating the energy minimizers using provably-convergence proximal gradient steps, the resulting architecture borrows the same basic structure as certain graph neural network layers that: (i) have been  fine-tuned to accommodate hypergraphs, and (ii) maintain the inductive bias infused by the original energy function.

\item The resulting framework, which we name \textbf{PhenomNN} for \textit{\underline{P}urposeful \underline{H}yper-\underline{E}dges i\underline{N} \underline{O}ptimization \underline{M}otivated \underline{N}eural \underline{N}etworks}, is applied to a multitude of hypergraph node classification benchmarks in Section \ref{sec:experiments}, achieving competitive or SOTA performance in each case. 
\end{itemize}

\section{Related Work} \label{sec:related_work}
\textbf{Hypergraph Expansions/Neural Networks}. Hypergraphs are frequently transformed into graphs by expansion methods including the clique and star expansions.  An extensive spectral analysis study of  of different hypergraph expansions is provided in \cite{agarwal2006higher}, but not from the vantage point of energy functions as is our focus. An alternative line expansion~\cite{Yang2020HypergraphLW} has also been proposed that can be viewed in some sense as a hybrid combination of clique and star expansions, although this involves the creation of additional nodes, and there may be scalability issues.  In terms of predictive models, previous spectral-based hypergraph neural networks are analogous to applying GNNs on clique expansions, including HGNN~\cite{hgnn}, HCHA~\cite{bai2021hypergraph}, H-GNNs~\cite{Zhang2022HypergraphCN}. Meanwhile, FastHyperGCN~\cite{hypergcn} and HyperGCN~\cite{hypergcn} reduce a hyperedge into a subgraph using Laplacian operators~\cite{chan2020generalizing}, which can be viewed as a modified form of clique expansion. HGAT~\cite{HyperGAT}, HNHN~\cite{dong2020hnhn}, HyperSAGE~\cite{arya2020hypersage}, UniGNN~\citep{UniGNN}, \cite{srinivasan2021learning}, Set-based models~\cite{chien2022you}, ~\cite{Heydari2022MessagePN}, ~\cite{Aponte2022AHN}, HEAT~\cite{Georgiev2022HEATHA} take into account hyperedge features and use a message-passing framework, which can be interpreted as GNNs applied to the star expansion graph.  And finally, \cite{wang2022equivariant} use gradient diffusion processes to motivate a broad class of hypergraph neural networks, although in the end there is not actually any specific energy function that is being minimized by the proposed model layers.

\textbf{Graph Neural Networks from Unfolded Optimization}. A variety of recent work has demonstrated that robust GNN architectures can be formed via graph propagation layers that mirror the unfolded descent iterations of a graph-regularized energy function \citep{chen2021graph,liu2021elastic,ma2020unified,pan2021a,yang2021graph,thatpaper,zhu2021interpreting,ahn2022descent}.  In doing so, the node embeddings at each layer can be viewed as increasingly refined approximations of an interpretable energy minimizer, that may be designed, for example, to mitigate GNN oversmoothing or perhaps inject robustness to spurious edges.  Furthermore, these learnable embeddings can be integrated within a bilevel optimization framework \citep{wang2016learning} for supervised training.  While at a high level we adopt a similar conceptual starting point, we nonetheless introduce non-trivial adaptations that are particular to the hypergraph domain, where this framework has not yet been extensively explored, and provide hypergraph-specific insights along the way.

\vspace*{-0.4cm}
\section{Hypergraph Background and Notation} \label{sec:notation}

A hypergraph can be viewed as a higher-order form of graph whereby edges can encompass more than two nodes.  Specifically, let $\mcG(\mcV,\mcE)$ denote a hypergraph, where $\calV$ is a set of $n = |\calV|$ vertices and $\calE$ is a set of $m= |\calE|$ hyperedges. In contrast to a traditional graph, each hyperedge $e_k \in \mcE$, can link an arbitrary number of nodes.  The corresponding hypergraph connectivity structure is conveniently represented in a binary incidence matrix $B \in \mathbb{R}^{n \times m}$, where $B_{ik}=1$ if node $v_i \in e_k$, otherwise  $B_{ik}=0$. We also use $D_H \in \mathbb{R}^{m \times m}$ to denote the degree matrix of the hypergraph, where $m_{e_k} \triangleq D_H[k,k]=\sum_{i}B_{ik}$. 

And finally, we define input features and embeddings for both nodes and hyperedges.  In this regard, $X \in \mathbb{R}^{n \times d_x}$ represents a matrix of $d_x$-dimensional initial/given node features, while $Y \in \mathbb{R}^{n \times d_y}$ refers to the corresponding node embeddings of size $d_y$ we seek to learn.  Analogously, $U \in \mathbb{R}^{n \times d_u}$ and $Z \in \mathbb{R}^{m \times d_z}$ are the initial edge features and learnable embeddings respectively. While here we have presented the most general form, we henceforth just assume $d=d_x=d_y=d_z=d_u$ for simplicity.

\vspace*{-0.3cm}
\section{A Family of Hypergraph Energy Functions} \label{sec:family}
\vspace*{-0.1cm}

Our goal is to pursue hypergraph-based energy functions whose minima produce embeddings that will ultimately be useful for downstream predictive tasks.  In this section, we first present an initial design of these functions followed by adaptations for handling the situation where no edge features $U$ are available.  We then show how in certain circumstances the proposed energy functions reduce to special cases that align with hypergraph star and clique expansions, before concluding with revised, simplified energy expressions informed by these considerations.
\vspace*{-0.3cm}
\subsection{Initial Energy Function Design and Motivation}
We begin with the general form 
\begin{align}
\label{eq:generalYU}
    \ell(Y,Z; \psi)=g_1(Y,X;\psi) + g_2(Z,U;\psi) + g_3(Y,Z,\mcG;\psi)
\end{align}
where $g_1(Y,X;\psi)$ and $g_2(Z,U;\psi)$ are non-structural  regularization factors over node and edge representations respectively, while $g_3(Y,Z,\mcG;\psi)$ explicitly incorporates hypergraph structure.  In all cases $\psi$ represents parameters that control the shape of the energy, with particular choices that should be clear from the context (note that these parameters need not all be shared across terms; however, we nonetheless lump them together for notational convenience). 

For the non-structural terms in (\ref{eq:generalYU}), a plausible design criteria is to adopt functions that favor embeddings (either node or edge) that are similar to the corresponding input features or some transformation thereof.  Hence we select 
\begin{eqnarray}
g_1(Y,X;\psi) &=& \sum_{i=1}^n \|y_i - f(x_i;W_x )\|_{2}^2  \nonumber \\
g_2(Z,U;\psi) &=& \sum_{k=1}^m \|z_k - f(u_k;W_u)\|_{2}^2, 
\end{eqnarray}
noting that both cases favor embeddings with minimal $\ell_2$ distance from  the trainable base predictor, and by extension, the initial features $\{X,U\}$. In practice, the function $f$ can be implemented as an MLP with node/edge weights $W_x$ and $W_u$ respectively.

Turning to $g_3(Y,Z,\mcG; \psi)$, our design is guided by the notion that: 
\begin{enumerate}[label=(\roman*)]
\item Both node and edge embeddings should be individually constrained to a shared subset of $\mathbb{R}^d$, e.g., consistent with most GNN architectures we may enforce non-negative embeddings;
\item Nodes sharing an edge should be similar when projected into an appropriate space, and;
\item Nodes within an edge set should have similar embeddings to the edge embedding, again, when suitably projected.
\end{enumerate}
With these desiderata in mind, we adopt

\begin{align}
\label{eq:g3}
&g_3(Y,Z,\mcG; \psi)=\sum_{i=1}^n \phi ( y_{i})+\sum_{k=1}^m \phi ( z_{i}) + \nonumber \\
    &\lambda_0\overbrace{\sum_{e_k \in \mathcal{E}}\sum_{i \in e_k}\sum_{j \in e_k}||y_iH_0-y_j||^2_2}^{(a)}   
        +\lambda_1\overbrace{\sum_{e_k \in \mathcal{E}}\sum_{i \in e_k}||y_iH_1-z_k||^2_2}^{(b)}
\end{align}

For the first terms we choose $\phi : \mathbb{R}^d \rightarrow \mathbb{R}^d_+$ defined as $\phi (p)\triangleq \sum_{j=1}^{d} \mathcal{I}_{\infty}[p_{i}<0]$, where $\mathcal{I}_{\infty}$ is an indicator function that assigns an infinite penalty to any $p_{i}<0$.  This ensures that all node and edge embedding must be non-negative to achieve finite energy.  Next, the term labeled (\textit{a}) in (\ref{eq:g3}) directly addresses criteria (ii).  We note that the summation is over both indices $i$ and $j$ so that the symmetric counterpart, where the roles of nodes $v_i$ and $v_j$ are switched, is effectively included in the summation.  And finally, criteria (iii) is handled by the last term, labeled (\textit{b}).  Here the node and edge embeddings play different roles and exhibit a natural asymmetry.\footnote{While we could consider adding an additional factor $||y_i -z_k H_2||^2_2$ to this term, we found that in practice it was not necessary.}  Incidentally, the projections $H_0$ and $H_1$ can be viewed as compatibility matrices, initially introduced for label or belief propagation~\cite{(ZooBP)Eswaran2017,(CAMLP)Yuto2016,Zhou2003} to provide additional flexibility to the metric in which entities are compared; for term (\textit{a}) $H_0$ facilitates the handling of nodes with potentially heterophily relationships, while for term (\textit{b}) $H_1$ accommodates the comparison of fundamentally different embedding types.

\vspace*{-0.3cm}
\subsection{Handling a Lack of Edge Features}

In some practical situations there may not be any initial hyperedge features $U$.  In such cases we could potentially modify $\ell(Y,Z;\psi)$ accordingly in multiple different ways.  First, and perhaps simplest, we can simply remove $g_2(Z,U;\psi)$ from (\ref{eq:generalYU}).  We will explore the consequences of this option further in Section \ref{sec:connect_to_expansions}.  But for tasks more related to hyperedge classification, it may be desirable to maintain this term for additional flexibility.  Hence as a second option, we could instead create pseudo features $\widetilde{U}$ with $\widetilde{u}_k = \mbox{AGG}\left[\{x_i | i \in e_k \} \right]$ for all $e_k \in \calE$ for some aggregation function $\mbox{AGG}$.  Or in a similar spirit, we could adopt $f(u_k; W_u) \equiv \mbox{AGG}\left[\{f(x_i;W_x) | i \in e_k \} \right]$ such that aggregation now takes place after the initial feature transformations.
\vspace*{-0.3cm}
\subsection{Analysis of Simplified Special Cases} \label{sec:connect_to_expansions}

Because most hypergraph benchmarks for node classification, and many real-world use cases, involve data devoid of hyperedge features, in this section we more closely examine simplifications of (\ref{eq:generalYU}) that arise when $g_2(Z,U;\psi)$ is removed.  For analysis purposes, it is useful to first introduce two representative hypergraph expansions, both of which can be viewed as converting the original hypergraph to a regular graph, which is tantamount to the assumption that edges in these expanded graphs involve only pairs of nodes.

\textbf{Clique Expansion.} For the \textit{clique expansion} \cite{784130}, we form the regular graph $\calG_C(\calV,\calE_C)$, where the node set $\calV$ remains unchanged while the edge set $\calE_C$ is such that, for all $e_k \in \calE$, we have that $\{v_i | i \in e_k \}$ forms a complete subgraph of $\calG_C$.  We define $L_C$, $A_C$, and $D_C$ as the corresponding Laplacian, adjacency matrix, and degree matrix of $\calG_C$ respectively.

\textbf{Star Expansion.} In contrast, the \textit{star expansion} \cite{784130} involves creating the bipartite graph $\mcG_{S}(\mcV_{S},\mcE_{S})$, with revised node set $\mcV_{S}=\{v_1, \ldots, v_{n+m} \}$ and edge set $\calE_S$ defined such that $\{v_i,v_{n+k}\} \in \calE_S$ iff $B_{ik} = 1$.  Conceptually, the resulting graph is formed with a new node associated with each hyperedge (from the original hypergraph), and an edge connecting every such new node to the original nodes within the corresponding hyperedges.  Additionally, $L_S = D_S-A_S$ is the revised Laplacian matrix, with $D_S$ and $A_S$ the degree and adjacent matrices of the star expansion graph.

\textbf{Unification.}  We now introduce simplifying assumptions to link the proposed energy with the Laplacians of clique and star expansions as follows:
% \vspace*{-1.5cm}
\begin{proposition}
\label{proposition:energy}
Suppose  $g_2(Z,U;\psi)$ is removed from (\ref{eq:generalYU}), $H_0 = H_1 = I$, and define $Z^* \triangleq D_H^{-T} B^T Y$.  It then follows that
\begin{eqnarray}
&& \hspace*{-0.8cm} \min_Z~\ell(Y,Z; \psi) ~~ = ~~ g_1(Y,X;\psi) + \sum_{i=1}^n \phi ( y_{i})  \\
% &+ & \hspace*{-0.3cm} \lambda_0 \sum_{e_k \in \mathcal{E}}\sum_{i \in e_k}\sum_{j \in e_k}||y_i -y_j||^2_2 + \lambda_1 \sum_{e_k \in \mathcal{E}}\sum_{i \in e_k}||y_i-z^*_k||^2_2 \nonumber \\
% & = &   g_1(Y,X;\psi) + \sum_{i=1}^n \phi ( y_{i}) \nonumber \\
&+ &  2\lambda_0 \mbox{tr}[Y^T L_{C} Y] + \lambda_1 \text{tr}\left(\left[\begin{array}{c}
         Y  \\
         Z^* 
    \end{array}\right]^TL_{S}\left[\begin{array}{c}
         Y  \\
         Z^*
    \end{array}\right] \right) \nonumber \\
    & & \hspace*{-0.8cm} = ~ g_1(Y,X;\psi) + \sum_{i=1}^n \phi ( y_{i})   \nonumber \\  &+ &2\lambda_0 \mbox{tr}[Y^T L_{C} Y] + \lambda_1 \mbox{tr}[Y^T \bar{L}_S Y],\nonumber 
\end{eqnarray}

where $\bar{L}_S \triangleq \bar{D}_S - \bar{A}_S$, with $\bar{A}_S \triangleq B D_H^{-1} B^T$ and $\bar{D}_S$ a diagonal matrix with nonzero elements formed as the corresponding row-sums of  $\bar{A}_S$. Moreover, if $\calG$ is $m_e$-uniform,\footnote{An $m_e$-uniform hypergraph is such that every hyperedge joins exactly $m_e$ nodes.  Hence a regular graph is by default a 2-uniform hypergraph.} then under the same assumptions
\begin{equation}
 \min_Z~\ell(Y,Z; \psi)  = g_1(Y,X;\psi) + \sum_{i=1}^n \phi ( y_{i}) + \beta \mbox{tr}[Y^T L_{C} Y],
\end{equation}
where $\beta \triangleq 2\lambda_0 + \tfrac{\lambda_1}{m_e}$.
\end{proposition}
All proofs are deferred to Appendix \ref{appendix:proofs}. This last result demonstrates that, under the stated assumptions, the graph-dependent portion of the original hypergraph energy, after optimizing away the influence of $Z$, can be reduced to a weighted quadratic penalty involving the graph Laplacian of the clique expansion.  Moreover, this factor further resolves as
\begin{equation}
  \mbox{tr}[Y^T L_{C} Y] = \frac{1}{2}\sum_{e_k \in \mcE} \sum_{i\in e_k}\sum_{j \in e_k}||y_i-y_j||^2_2.
\end{equation}
Of course in more general settings, for example when $H_0 \neq H_1 \neq I$, or when $\phi(p) \neq \sum_{j=1}^{d} \mathcal{I}_{\infty}[p_{i}<0]$,  this equivalence will \textit{not} generally hold.

\subsection{Revised Hypergraph Energy Functions}
The analysis from the previous sections motivates two practical, revised forms of our original energy from (\ref{eq:generalYU}), which we will later use for all of our empirical evaluations.  For convenience, we define
\begin{equation}
\ell(Y;\psi) \triangleq \ell(Y, Z = Z^*;\psi ).
\end{equation}
Then the first, more general variant, we adopt is  
\begin{align}
\label{eq:PhenomNN-general-matrix}
& \ell(Y ;\psi=\{W,H_0,H_1\}) \nonumber \\
        &=~|| Y-f(X;W)||_{\mathcal{F}}^2 +\sum_{i} \phi ( y_{i})  +\nonumber\\
        &\lambda_0\overbrace{\text{tr}\Bigg[(YH_0)^TD_{C}YH_0  -2(YH_0)^TA_{C}Y+  Y^TD_{C}Y\Bigg]}^{(a)}  + \nonumber\\
        &\lambda_1\overbrace{\text{tr}\Bigg[(YH_1)^T \bar{D}_SYH_1-2(YH_1)^TBZ^*+Z^{*T}D_{H}Z^*\Bigg]}^{(b)},
\end{align}
where $\bar{D}_S$ is defined as in Proposition \ref{proposition:energy}. Moreover, to ease later exposition, we have overloaded the definition of $f$ such that $\| Y-f(X;W)\|_{\mathcal{F}}^2 \equiv \sum_{i=1}^n \|y_i - f(x_i; W)\|_2^2$.   And secondly, as a less complex alternative we have
\begin{align}
\label{eq:PhenomNN-matrix}
        & \ell(Y ;\psi=\{W,I,I\})=  \\ 
        & || Y-f(X;W)||_{\mathcal{F}}^2 + \sum_{i} \phi ( y_{i}) 
        +\text{tr}[Y^T(\lambda_0L_{C}+\lambda_1 \bar{L}_{S})Y]. \nonumber
\end{align}

\section{Hypergraph Node Classification via Bilevel Optimization} \label{sec:optimization}

We now demonstrate how the optimal embeddings obtained by minimizing the energy functions from the previous section can be applied to our ultimate goal of hypergraph node classification.  For this purpose, define
\begin{equation}
Y^*(\psi) = \arg\min_{Y} \ell(Y; \psi), 
\end{equation}
noting that the solution depends explicitly on the parameters $\psi$ governing the shape of the energy.  We may then consider treating $Y^*(\psi)$, which is obtainable from the above optimization process, as features to be applied to a discriminative node classification loss $\calD$ that can be subsequently minimized via a second, meta-level optimization step.\footnote{Because our emphasis is hypergraph node classification, we will not explicitly use any analogous hyperedge embeddings for the meta-level optimization; however, they nonetheless still play a vital role given that they are co-adapted with the node embeddings during the lower-level optimization per the discussion from the previous section.}  In aggregate we arrive at the \textit{bilevel} optimization problem
\begin{align}
    \label{eq:bileveloptim}
    \ell(\theta,\psi) \triangleq \sum_{i=1}^{n'}\mathcal{D}(h[y_i^*(\psi);\theta],\tau_i),
\end{align}
where $\calD$ is chosen as an classification-friendly cross-entropy function, $y_i^*(\psi)$ is the $i$-th row of $Y^*(\psi)$, and $\tau_i \in \mathbb{R}^c$ is the ground-truth label of node $i$ to be approximated by some differentiable node-wise function $h : \mathbb{R}^{d} \rightarrow \mathbb{R}^{c}$ with trainable parameters $\theta$.  We have also implicitly assumed that the first $n'$ nodes of $\calG$ are labeled.  Intuitively, (\ref{eq:bileveloptim}) involves training a classifier $h$, with input features $y_i^*(\psi)$, to predict labels $\tau_i$.

At this point, assuming $\partial Y^*(\psi)/\partial \psi$ is somehow computable, then $\ell(\psi,\theta)$ can be efficiently trained over \textit{all} parameters, including $\psi$ from the lower level optimization.  However, directly computing $\partial Y^*(\psi)/\partial \psi$ is not generally feasible.  Instead, in the remainder of this section we will derive approximate embeddings $\hat{Y}(\psi) \approx Y^*(\psi)$ whereby $\partial \hat{Y}(\psi)/\partial \psi$ can be computed efficiently.  And as will be assessed in greater detail later, the computational steps we derive to produce  $\hat{Y}(\psi)$ will mirror the layers of canonical graph neural network architectures.  It is because of this association that we refer to our overall model as \textbf{PhenomNN}, for \textit{\underline{P}urposeful \underline{H}yper-\underline{E}dges i\underline{N} \underline{O}ptimization \underline{M}otivated \underline{N}eural \underline{N}etworks} as mentioned in the introduction.

\subsection{Deriving Proximal Gradient Descent Steps}

To efficiently deploy proximal gradient descent (PGD) \cite{(Proximal)Parikh2014}, we first must split our loss into a smooth, differentiable part, and a non-smooth but separable part.  Hence we adopt the decomposition
\begin{align}
   \ell(Y;\psi) = \bar{\ell}(Y;\psi) + \sum_{i} \phi ( y_{i}), \label{eq:energy_revised}
\end{align}
where $\bar{\ell}(Y;\psi)$ is defined by exclusion upon examining the original form of $\ell(Y;\psi)$. The relevant proximal operator is 
\begin{eqnarray}
    \bprox_{\phi}(V) & \triangleq & \arg \min_{Y} \frac{1}{2} ||V-  Y||_\calF^2 + \sum_i \phi(y_i) \nonumber \\
    & = & \text{max}(0, V),
    \label{eq:prox_operator}
\end{eqnarray}
where the max operator is assumed to apply elementwise.  Subsequent PGD iterations for minimizing (\ref{eq:energy_revised}) are then computed as 
\begin{align}
    \bar{Y}^{(t+1)} & = Y^{(t)} - \alpha \Omega \nabla_{Y^{(t)}}\bar{\ell}(Y^{(t)};\psi) \label{eq:before_prox} \\
    Y^{(t+1)} & = \text{max}(0, \bar{Y}^{(t+1)}),  \label{eq:after_prox}
\end{align}
% For $\phi (y)=\sum_{i=1}^{d} \mathcal{I}_{\infty}[y_{i}<0]$, the corresponding proximal operator becomes $\bprox_{\phi}(v)=\text{ReLU}(v)=\text{max}(\textbf{0}, v)$, 
where $\alpha$ is a step-size parameter and $\Omega$ is a positive-definite pre-conditioner to be defined later.  Incidentally, as will become apparent shortly, (\ref{eq:before_prox}) will occupy the role of a pre-activation hypergraph neural network layer, while (\ref{eq:after_prox}) provides a ReLU nonlinearity.  A related association was previously noted within the context of traditional GNNs \cite{yang2021graph}.  We now examine two different choices for $\Omega$ and $\psi$ that correspond with the general form from (\ref{eq:PhenomNN-general-matrix}) and the simplified alternative from (\ref{eq:PhenomNN-matrix}).

\textbf{General Form}. To compute \eqref{eq:before_prox}, we consider term (\textit{a}) and (\textit{b}) from \eqref{eq:PhenomNN-general-matrix} separately.  Beginning with (\textit{a}), the corresponding gradient is 
\begin{align}
\label{eq:H-clique-gradient}
    2D_{C}Y-2\Tilde{Y}_{C},
\end{align}
where $\Tilde{Y}_{C} \triangleq A_{C}Y(H_0+H_0^T)-D_{C}YH_0H_0^T$. Similarly, for $(\textit{b})$ the gradient is given by
\begin{align}
\label{eq:gradient-gamma}
    2BD_H^{-1}D_H(BD_H^{-1})^{T}Y-2\Tilde{Y}_{S},
\end{align}
where $\Tilde{Y}_{S} \triangleq (B(BD_H^{-1})^{T}YH_1^T+BD_H^{-1}B^TYH_1)-\bar{D}_{S}YH_1H_1^T$. Additionally, given that $BD_H^{-1}D_H(BD_H^{-1})^{T}=B(BD_H^{-1})^{T}=BD_H^{-1}B^T=\bar{A}_{S}$, we can reduce \eqref{eq:gradient-gamma} to 
\begin{align}
    2\bar{A}_{S}Y-2\Tilde{Y}_{S},
\end{align}
since now $\Tilde{Y}_{S}=\bar{A}_{S}Y(H_1+H_1^T)-\bar{D}_{S}YH_1H_1^T$. 
Combining terms, the gradient for $\bar{\ell}(Y;\psi)$ is 
\begin{align}
\frac{\bar{\ell}(Y;\psi)}{\partial Y} =2\lambda_0(D_{C}Y-\Tilde{Y}_{C})+ 2\lambda_1(\bar{A}_{S}Y-\Tilde{Y}_{S}) \nonumber\\
+ 2Y - 2 f\left(X ; W  \right),
\end{align}
and \eqref{eq:before_prox} becomes
\begin{align} 
\label{eq:third_H-star}
\bar{Y}^{(t+1)} = Y^{(t)} - \alpha\Bigg[ \lambda_0(D_{C}Y^{(t)}-\Tilde{Y}_{C}^{(t)}) \nonumber \\ + 
\lambda_1(\bar{A}_{S}Y^{(t)}-\Tilde{Y}_{S}^{(t)}) + Y^{(t)} - f\left(X ; W  \right) \Bigg], 
\end{align}
where $\alpha/2$ is the step size. The coefficient $\bar{\Omega}$ before $Y^{(t)}$ is
\begin{equation}
    \bar{\Omega}\triangleq \lambda_0 D_{C}+ \lambda_1 \bar{A}_{S} + I.
\end{equation}
Applying Jacobi preconditioning~\cite{Axelsson1996} often aids convergence by helping to normalize the scales across different dimensions. One natural candidate for the preconditioner is $\left( \mbox{diag}[\bar{\Omega}] \right)^{-1}$; however, we use the more spartan $\Omega = \tilde{D} ^{-1}$ where $\tilde{D} \triangleq \lambda_0 D_{C}+ \lambda_1 \bar{D}_{S} + I$. After rescaling and applying \eqref{eq:after_prox}, the composite PhenomNN update is given by
\begin{align} 
\label{eq:PhenomNN-general-updating}
Y^{(t+1)} &= \text{ReLU}\Bigg((1-\alpha)Y^{(t)} + \alpha \tilde{D} ^{-1}\Big[ f\left(X ; W \right) \nonumber\\&+ \lambda_0\Tilde{Y}_{C}^{(t)}+ 
\lambda_1(\bar{L}_{S}Y^{(t)}+\Tilde{Y}_{S}^{(t)})  \Big]\Bigg),
\end{align}
where $\bar{L}_{S}=\bar{D}_{S}-\bar{A}_{S}$ as in Proposition \ref{proposition:energy}.  This respresents the general form of PhenomNN.

\textbf{Simplified Alternative. } Regarding the simplified energy from \eqref{eq:PhenomNN-matrix}, the relevant gradient is 
\begin{align}
\frac{\partial \bar{\ell}(Y;\psi={W,I,I})}{\partial Y} = 2(\lambda_0L_{C}+\lambda_1\bar{L}_{S}) Y + \nonumber\\
2Y - 2 f\left(X ; W  \right),
\end{align}
leading to the revised update
\begin{align} \label{eq:PhenomNN_basic_grad_step}
\bar{Y}^{(t+1)} = Y^{(t)} - \alpha\left[ \Tilde{\Omega} Y^{(t)} - f\left(X ; W  \right) \right], \nonumber\\
\mbox{with}~~\Tilde{\Omega} \triangleq \lambda_0L_{C}+\lambda_1\bar{L}_{S}  + I
\end{align}
and step size $\alpha/2$ as before.  And again, we can apply preconditioning, in this case rescaling each gradient step by 
$\Omega=\left( \mbox{diag}[\Tilde{\Omega}] \right)^{-1} = \left(\lambda_0D_{C}+\lambda_1\bar{D}_{S} + I\right)^{-1} =\tilde{D}^{-1}$. So the final/composite update formula, including \eqref{eq:after_prox}, becomes 
\begin{align} \label{eq:PhenomNN-updating}
Y^{(t+1)} = \text{ReLU}\Big((1-\alpha) Y^{(t)} + \nonumber\\
\alpha \tilde{D}^{-1} \left[ (\lambda_0A_{C}+\lambda_1\bar{A}_{S}) Y^{(t)} + f\left(X ; W  \right) \right]\Big).
\end{align}
We henceforth refer to this variant as  \textbf{$\text{PhenomNN}_{\text{simple}}$}.

\subsection{Overall Algorithm}
The overall algorithm for PhenomNN is demonstrated in Algorithm \ref{alg:hypergraph}.
\begin{algorithm}[h]
   \caption{PhenomNN Algorithm for Hypergraph Node Classification.}
   \label{alg:hypergraph}
\begin{algorithmic}
   \STATE {{\bfseries Input}: Hypergraph incidence matrix $B$, node features $X$, number of layers $T$, training epochs $E$, and node labels $\tau = \{\tau_i\}$.} 
\FOR{$e=0$ to $E-1$}
  \STATE{Set initial projection $Y^{(0)}=f(X;W)$, where $f$ is the trainable base model.}
  \FOR{$t=0$ to $T-1$}
    \STATE{$Y^{(t+1)} = Update(Y^{(t)})$, where $Update$ is computed via \eqref{eq:PhenomNN-general-updating} for PhenomNN or \eqref{eq:PhenomNN-updating} for PhenomNN$_\text{simple}$.}
  \ENDFOR
  \STATE{Compute loss $\ell(\theta,\psi) = \sum_i \mathcal{D}(h[y_i^{(T)};\theta],\tau_i)$ from \eqref{eq:bileveloptim}, where $\psi=\{W,H_0,H_1\}$ for PhenomNN and $\psi=\{W,I,I\}$ for PhenomNN$_\text{simple}$, noting that each $y_i^{(T)}$ is a trainable function of $\psi$ by design.}
  \STATE{Backpropagate over all parameters $\psi, \theta$ using optimizer (Adam, SGD, etc.)}
\ENDFOR

\end{algorithmic}
\end{algorithm}

\subsection{Convergence Analysis}

We now consider the convergence of the iterations \eqref{eq:PhenomNN-general-updating} and \eqref{eq:PhenomNN-updating} introduced in the previous section.  First, for the more general form we have the following:

\begin{proposition}
\label{proposition:gradientconverge}
The PhenomNN updates from \eqref{eq:PhenomNN-general-updating} are guaranteed to monotonically converge to the unique global minimum of $\ell(Y;\psi)$ on the condition that
\begin{align}
    \alpha < \frac{1 + \lambda_0 d_{C\text{min}} + \lambda_1d_{S\text{min}}}{1 + \lambda_0 d_{C\text{min}} + \sigma_{\text{max}}},
\end{align}
where $d_{C\text{min}}$ is the minimum diagonal element of $I \otimes D_{C}$, $d_{S\text{min}}$ is the minimum diagonal element of $I \otimes \bar{D}_{S}$
and $\sigma_{\text{max}} \text{ is the max eigenvalue of }(Q - P + \lambda_1I\otimes \bar{A}_{S})$ with
\begin{align}
    &Q \triangleq \lambda_0H_0^TH_0\otimes D_{C} + \lambda_1H_1^TH_1\otimes \bar{D}_{S}, \\
    &P \triangleq \lambda_0(H_0+H_0^T)\otimes A_{C} +\lambda_1(H_1+H_1^T)\otimes \bar{A}_{S}.
\end{align}
\end{proposition}

And for the restricted case where $\psi=\{W,I,I\}$,  the convergence conditions simplify as follows:
\begin{corollary}
\label{corollary:gradientconverge-simple} 
The $\text{PhenomNN}_{\text{simple}}$ updates from \eqref{eq:PhenomNN-updating} are guaranteed to monotonically converge to the unique global minimum of $\ell(Y;\psi=\{W,I,I\})$ on the condition that
\begin{align}
       \alpha < \frac{1+\lambda_0d_{Cmin}+\lambda_1d_{Smin}}{1+\lambda_0d_{Cmin}+\lambda_1d_{Smin} -\sigma_{min}},
\end{align}
where $\sigma_{min}$ is the min eigenvalue of $(\lambda_0A_{C}+\lambda_1\bar{A}_{S})$.
\end{corollary}

\subsection{Complexity Analysis}

Analytically, PhenomNN$_\text{simple}$ has a time complexity given by $O(|\mathcal{E}|Td+|\mathcal{V}|P d^2)$, where $|\mathcal{E}|$ is edge number, $|\mathcal{V}|$ is the node number, $T$ is the number of layers/iterations, $d$ is the hidden size, and $P$ is the number of MLP layers in $f(\cdot ;W)$. In contrast, for PhenomNN this complexity increases to $O(|\mathcal{E}|Td+|\mathcal{V}|(T+P)d^2)$, which is roughly the same as a standard GCN model. In fact, the widely-used graph convolution networks (GCN) \cite{kipf2016semi} have equivalent complexity to PhenomNN up to the factor of $P$ which is generally small (e.g., $P=1$ for PhenomNN in our experiments, while for a GCN $P = 0$). In this way then, PhenomNN$_\text{simple}$ is actually somewhat cheaper than a GCN when $T > P$. Additionally, we include complementary empirical results related to time and space complexity in Section \ref{sec:experiments}.

\section{Connections with Existing GNN Layers} \label{sec:connections}

As mentioned in Section \ref{sec:connect_to_expansions}, the clique and star expansions can be invoked to transform hypergraphs into homogeneous and bipartite graphs respectively (where the latter is a special case of a heterogeneous graph).  In this section we examine how the layer-wise structure of two of the most popular GNN models, namely GCN~\cite{kipf2016semi} mentioned previously, and relational graph convolution networks (RGCN) \cite{schlichtkrull2018modeling}, relate to PhenomNN and simplifications thereof.

% The clique expansion and star expansion transforms hypergraphs into homogeneous graphs or bipartite graphs (a special case of heterogeneous graphs). In this section we show that they can be generalized altogether in the energy of hypergraphs.
\subsection{Homogeneous Graphs and GCN}
Using the so-called message-passing form of expression, the embedding update for the $i$-th node of the $t$-th GCN layer can be written as 
\begin{align}
    \label{eq:gcn-messagepassing}
    y_i^{(t+1)}=\sigma\left(\sum_{j \in \mathcal{N}_i}\frac{1}{c_{ij}}W^{(t)}y_j^{(t)}\right)
\end{align}
where $\sigma$ is an activation function like ReLU, $W^{(t)}$ are weights, $c_{ij} \triangleq \sqrt{|\mathcal{N}_i||\mathcal{N}_j|}$ and $\mathcal{N}_i$ refers to the set of neighboring nodes in some input graph (note also that the graph could have self-loops in which case $i \in \mathcal{N}_i$).  Interestingly, follow-up work \cite{ma2020unified,pan2021a,yang2021graph,thatpaper,zhu2021interpreting} has demonstrated that
this same basic layer-wise structure can be closely linked to iterative steps designed to minimize the energy
\begin{align}
\label{eq:energy_functionTWIRLS}
    \ell(Y)=|| Y-f(X;W)||_{\mathcal{F}}^2 + \lambda\text{tr}[Y^TLY],
\end{align}
where $f$ is defined as before and $L$ is the assumed graph Laplacian matrix. One way to see this is to examine a preconditioned gradient step along (\ref{eq:energy_functionTWIRLS}), which can be expressed as
\begin{align} \label{eq:twirls_update}
Y^{(t+1)} = (1-\alpha) Y^{(t)} + \alpha \widetilde{D}_0^{-1} \left[ \lambda A Y^{(t)} + f\left(X ; W  \right) \right],
\end{align}
with preconditioner $\tilde{D}_0^{-1} = \left(\lambda D + I\right)^{-1}$, step-size parameter $\alpha$, graph adjacency matrix $A$, and corresponding degree matrix $D$.  Moreover, for a single node $i$, \eqref{eq:twirls_update} can be reduced to
\begin{align}
    \label{eq:TWIRLS-messagepassing}
    y_i^{(t+1)} =  \left(\sum_{j \in \mcN_{i}} \frac{1}{\tilde{c}_i} y_j^{(t)} \right) + \tilde{f}_i(x_i;W), 
\end{align}
where $\tilde{c}_i$ is a scaling constant dependent on $\lambda$, the gradient step-size, and the preconditioner, while $\tilde{f}_i$ is merely $f$ similarly rescaled.  If we add an additional penalty $\phi$ and subsequent proximal operator step to introduce a non-linearity, then this result is very similar to (\ref{eq:gcn-messagepassing}), although without the weight matrix directly on each $y_j^{(t)}$ but with an added skip connection to the input layer.

Importantly for our purposes though, if the input graph is chosen to be a hypergraph clique expansion, and we set $D = D_C$, $A = A_C$, $\lambda = \lambda_0$, and $\lambda_1 = 0$, then we arrive at a special case of $\text{PhenomNN}_{\text{simple}}$ from \eqref{eq:PhenomNN-updating}.  Of course one might not naturally conceive of the more generalized form that leads to $\text{PhenomNN}_{\text{simple}}$, and by extension PhenomNN, without the interpretable grounding of the underlying hypergraph energy functions involved.

\begin{table*}[htbp]
\def\p{ ± } 
\centering

\setlength\tabcolsep{2.5pt}
\caption{Results on datasets from \cite{Zhang2022HypergraphCN}: Mean accuracy (\%)  ±  standard deviation results over 10 train-test splits. Boldfaced letters are used to indicate the best mean accuracy and underline for the second. "-" means not reported in their paper so in average ranking we just average over the ones that are available.  OOM indicates out-of-memory.}
\scriptsize
    \begin{tabular}{c|ccccccc|c}
        \toprule 
              \multicolumn{1}{c}{}&\multicolumn{1}{c}{\begin{tabular}[c]{@{}c@{}}Cora\\ (co-authorship)\end{tabular}}  & \multicolumn{1}{c}{\begin{tabular}[c]{@{}c@{}}DBLP\\ (co-authorship)\end{tabular}} & \multicolumn{1}{c}{\begin{tabular}[c]{@{}c@{}}Cora\\ (co-citation)\end{tabular}} & \multicolumn{1}{c}{\begin{tabular}[c]{@{}c@{}}Pubmed\\ (co-citation)\end{tabular}} & \multicolumn{1}{c}{\begin{tabular}[c]{@{}c@{}}Citeseer\\ (co-citation)\end{tabular}}& \multicolumn{1}{c}{\begin{tabular}[c]{@{}c@{}}NTU2012\\ (both features)\end{tabular}}&\multicolumn{1}{c}{\begin{tabular}[c]{@{}c@{}}ModelNet40\\ (both features)\end{tabular}} & Avg Ranking \\
        \midrule
             MLP+HLR  & 59.8\p4.7 & 63.6\p4.7 & 61.0\p4.1 & 64.7\p3.1 & 56.1\p2.6&- & -& 13.6 \\
             FastHyperGCN  & 61.1\p8.2 & 68.1\p9.6 & 61.3\p10.3 & 65.7\p11.1 & 56.2\p8.1&- & - & 12.4\\
             HyperGCN  & 63.9\p7.3 & 70.9\p8.3 & 62.5\p9.7& 68.3\p9.5 & 57.3\p7.3&- & -& 10.8 \\
            HGNN  & 63.2\p3.1 & 68.1\p9.6 & 70.9\p2.9& 66.8\p3.7 & 56.7\p3.8  &83.54 ± 0.50 & 97.15 ± 0.14& 9.4 \\
             HNHN  & 64.0\p 2.4 & 84.4\p 0.3& 41.6\p 3.1 & 41.9\p4.7 & 33.6\p 2.1 &- & -& 13.0 \\
             HGAT  & 65.4 ± 1.5 & OOM &52.2 ± 3.5 & 46.3 ± 0.5 & 38.3 ± 1.5 &84.05 ± 0.36 & 96.44 ± 0.15& 12.0\\
             HyperSAGE  &  72.4\p1.6 & 77.4\p3.8 & 69.3\p2.7 &72.9\p1.3 & 61.8\p2.3&- & - & 8.6\\
             UniGNN  & 75.3\p1.2 & 88.8\p0.2 & 70.1\p1.4 & 74.4\p 1.0& 63.6\p 1.3&84.45 ± 0.40 & 96.69 ± 0.07& 6.0 \\
        
        \midrule
            H-ChebNet   & 70.6 ± 2.1 & 87.9 ± 0.24 &  69.7 ± 2.0 & 74.3 ± 1.5 & 63.5 ± 1.3 &83.16 ± 0.46 & 96.95 ± 0.09 & 8.0\\
            H-APPNP  & 76.4 ± 0.8 & 89.4 ± 0.18 & 70.9 ± 0.7 &  75.3 ± 1.1 & 64.5 ± 1.4 &83.57 ± 0.42 & 97.20 ± 0.14 & 4.6\\
            H-SSGC  & 72.0 ± 1.2 & 88.6 ± 0.16 &68.8 ± 2.1  & 74.5 ± 1.3 & 60.5 ± 1.7 &84.13 ± 0.34 & 97.07 ± 0.07 & 7.6 \\
            H-GCN  & 74.8\p0.9 & 89.0\p 0.19 & 69.5\p 2.0 & 75.4\p1.2 & 62.7\p1.2 &84.45 ± 0.40 & 97.28 ± 0.15& 5.4 \\
            H-GCNII  & 76.2 ± 1.0 & \underline{89.8 ± 0.20} & 72.5 ± 1.2
            & 75.8 ± 1.1 & 64.5 ± 1.0 & 85.17 ± 0.36 & 97.75 ± 0.07& 3.0 \\
                    \midrule

    $\text{PhenomNN}_{\text{simple}}$  & \textbf{77.62  ±  1.30} & 89.74  ±  0.16 &72.81  ±  1.67
            & \underline{76.20  ±  1.41} &\underline{ 65.07  ±  1.08} &\underline{85.39 ± 0.40} & \textbf{97.83 ± 0.09}& \underline{1.9}\\

PhenomNN  & \underline{77.11  ±  0.45} & \textbf{89.81  ±  0.05} &\textbf{73.09  ±  0.65}
            & \textbf{78.12  ±  0.24} & \textbf{65.77  ±  0.45} & \textbf{85.40 ± 0.42} & \underline{97.77 ± 0.11}& \textbf{1.3}\\

        \hline
        \bottomrule

    \end{tabular}
    
\label{tab:icmlworkshopdataset} 
\end{table*}
\subsection{Heterogeneous Graphs and RGCN} \label{sec:rgcn}
For heterogeneous graphs applied to RGCN, the analogous message-passing update for the $i$-th node in the $t$-th layer is given by
\begin{align}
    \label{eq:RGCN-messagepassing}
    y_{i}^{(t+1)} =\sigma\left(\sum_{r \in \mathcal{R}}\sum_{j \in \mathcal{N}_{i}^{r}}\frac{1}{c_{i,r}}y_j^{(t)}W_{r}^{(t)} + y_{i}^{(t)}W_{0}^{(t)} \right),
\end{align}
where $\mcR$ is the set of edge types in a heterogeneous input graph, $\mcN_i^r$ is the set of neighbors with edge type $r$, $c_{i,r} \triangleq |\mcN_i^r|$, and $W_r^{(t)}$ and $W_0^{(t)}$ are weight/projection matrices.  In this context, the RGCN input could conceivably be chosen as the bipartite graph produced by a given star expansion (e.g., such a graph could be assigned the edge types ``hypergraph node belongs to hyperedge" and ``hyperedge belongs to hypergraph node").

For comparison purposes, we can also re-express our general PhenomNN model from \eqref{eq:PhenomNN-general-updating}, in the node-wise message-passing form
\begin{align} \label{eq:mp_PhenomNN}
    y_i^{(t+1)}  =\sigma \left(\sum_{j \in \mcN_i^C}y_{j}^{(t)}W_{ij}^{(t)} + y_{i}^{(t)}W_{i}^{(t)} + \alpha \tilde{D} ^{-1}_{ii} f\left(x_i ; W \right) \right),
\end{align}

where $\calN_i^C$ are neighbors in the clique (not star) expansion graph (more on this below) and the weight matrices are characterized by the special energy-function-dependent forms
\begin{align}
    W_{ij}^{(t)} &\triangleq \alpha \tilde{D} ^{-1}_{ii}\Big[ \lambda_0 A_C[i,j](H_0+H_0^T) + \nonumber \\
    &\lambda_1 \bar{A}_S[i,j](H_1+H_1^T - I)\Big],\\
    W_{i}^{(t)}&\triangleq(1-\alpha)I-\alpha \tilde{D} ^{-1}_{ii}\Big[\lambda_0 D_C[i,i] H_0H_0^T+\nonumber\\
    &\lambda_1 \bar{D}_S[i,i](H_1H_1^T-I) \Big]. 
    % &a_{Cij}\triangleq A_C[i,j], a_{Sij}\triangleq \bar{A}_S[i,j],  \\
    % &d_{Ci}\triangleq D_C[i,i], d_{Si}\triangleq \bar{D}_S[i,i].
\end{align}
While the basic structures of \eqref{eq:RGCN-messagepassing} and \eqref{eq:mp_PhenomNN} are similar, there are several key differences:
\begin{itemize}
\item When RGCN is applied to the star expansion, neighbors are defined by the resulting bipartite graph, and nodes in the original hypergraph do not directly pass messages to each other.  In contrast, because within PhenomNN we have optimized away the hyperedge embeddings, the implicit graph that dictates neighborhood structure is actually the \textit{clique expansion graph} as reflected in \eqref{eq:mp_PhenomNN}.
\item The PhenomNN projection matrices have special structure infused from the energy function and optimization over the edge embeddings.  As such, unlike RGCN node $i$ receives messages from its connected neighbors and itself, with  projection matrices $W_{ij}^{(t)}$ and $W_{i}^{(t)}$ that can vary from node to node and edge to edge.  In contrast, RGCN has layer-wise (or analogously iteration-wise) dependent weights.
\item PhenomNN has an additional weighted skip connection from the input base model $f\left(x_i ; W \right)$.  While of course RGCN could also be equipped with a similar term, this would be accomplished in a post-hoc fashion, and not tethered to an underlying energy function.
\end{itemize}

\begin{table*}[htbp]
\centering
\setlength{\tabcolsep}{2.5pt}
\caption{Results using the benchmarks from \cite{chien2022you}: Mean accuracy (\%)  ±  standard deviation. The number behind Walmart and House is the feature noise standard deviation for each dataset, and for HAN$^*$, additional preprocessing of each dataset is required (see \cite{chien2022you} for more details). Boldfaced letters are used to indicate the best mean accuracy and underline is for the second. OOM indicates out-of-memory.} % % OOM indicates out-of-memory. For computing rank, OOM is put in the last place.
\label{tab:iclr22dataset} 
\scriptsize
\begin{tabular}{@{}c|cccccccc|c@{}}
\toprule

\multicolumn{1}{l}{} &
  $\text{NTU2012}^{*}$ &
  $\text{ModelNet40}^{*}$ &
  Yelp &
  House(1) &
  Walmart(1) &
  House(0.6) &
  Walmart(0.6) &
  20Newsgroups
  & Avg Ranking
   \\ \midrule

MLP &
  85.52 ± 1.49 &
  96.14 ± 0.36 &
  31.96 ± 0.44 &
  67.93 ± 2.33 &
  45.51 ± 0.24 &
  81.53 ± 2.26 &
  63.28 ± 0.37 &
  \underline{81.42 ± 0.49}& 6.9
   \\
CEGCN &
  81.52 ± 1.43 &
  89.92 ± 0.46 &
  OOM &
  62.80 ± 2.61 &
  54.44 ± 0.24 &
  64.36 ± 2.41 &
  59.78 ± 0.32 & OOM & 11.5
   \\
CEGAT &
  82.21 ± 1.23 &
  92.52 ± 0.39 &
  OOM &
  69.09 ± 3.00 &
  51.14 ± 0.56 &
  77.25 ± 2.53 &
  59.47 ± 1.05 & OOM& 10.4
   \\
HNHN &
  89.11 ± 1.44 &
  97.84 ± 0.25 &
  31.65 ± 0.44 &
  67.80 ± 2.59 &
  47.18 ± 0.35 &
  78.78 ± 1.88 &
  65.80 ± 0.39 & 81.35 ± 0.61& 7.1
   \\
HGNN &
  87.72 ± 1.35 &
  95.44 ± 0.33 &
  \underline{33.04 ± 0.62} &
  61.39 ± 2.96 &
  62.00 ± 0.24 &
  66.16 ± 1.80 &
  77.72 ± 0.21 &80.33 ± 0.42& 7.8
   \\
HCHA &
  87.48 ± 1.87 &
  94.48 ± 0.28 &
  30.99 ± 0.72 &
  61.36 ± 2.53 &
  62.45 ± 0.26 &
  67.91 ± 2.26 &
  77.12 ± 0.26 & 80.33 ± 0.80& 8.8
   \\
HyperGCN &
  56.36 ± 4.86 &
  75.89 ± 5.26 &
  29.42 ± 1.54 &
  48.31 ± 2.93 &
  44.74 ± 2.81 &
  78.22 ± 2.46 &
  55.31 ± 0.30 & 81.05 ± 0.59& 12
   \\
UniGCNII &
  89.30 ± 1.33 &
  98.07 ± 0.23 &
  31.70 ± 0.52 &
  67.25 ± 2.57 &
  54.45 ± 0.37 &
  80.65 ± 1.96 &
  72.08 ± 0.28 & 81.12 ± 0.67& 6.2
   \\
HAN (full batch)$^*$ &
  83.58 ± 1.46 &
  94.04 ± 0.41 &
  OOM &
  \underline{71.05 ± 2.26} &
  OOM &
  83.27 ± 1.62 &
  OOM & OOM& 9.6
   \\
HAN (mini batch)$^*$ &
  80.77 ± 2.36 &
  91.52 ± 0.96 &
  26.05 ± 1.37 &
  62.00 ± 9.06 &
  48.57 ± 1.04 &
  82.04 ± 2.68 &
  63.10 ± 0.96 & 79.72 ± 0.62& 10.4
  \\
  AllDeepSets &
  88.09 ± 1.52 &
  96.98 ± 0.26 &
  30.36 ± 1.57 &
  67.82 ± 2.40 &
  \underline{64.55 ± 0.33} &
  80.70 ± 1.59 &
  \textbf{78.46 ± 0.26} & 81.06 ± 0.54 &5.6
   \\
  AllSetTransformer &
  88.69 ± 1.24 &
  98.20 ± 0.20 &
  \textbf{36.89 ± 0.51} &
  69.33 ± 2.20 &
  \textbf{65.46 ± 0.25} &
  83.14 ± 1.92 &
  \textbf{78.46 ± 0.40} &81.38± 0.58
  &\underline{3.1}
   \\

  \hline
      $\text{PhenomNN}_{\text{simple}}$ &
  \textbf{91.03 ± 1.04} &
  \textbf{98.66 ± 0.20} &
32.26 ± 0.40&
   \textbf{71.77 ± 1.68} &
  \textbf{64.11  ±  0.49} &
 \textbf{86.96 ± 1.33} &
 \textbf{78.46 ± 0.32} &
\textbf{81.74 ± 0.52} &
\textbf{1.6}
   \\
      PhenomNN &
  \underline{90.62 ± 1.88} &
  \underline{98.61 ± 0.17} &
31.92 ± 0.36 &
  70.71  ±  2.35  &
  62.98  ±  1.36  &
  \underline{85.28  ±  2.30} &
 \underline{78.26 ± 0.26} &
81.41 ± 0.49 &
\underline{3.1}
\\

  \bottomrule
  
\end{tabular}

\end{table*}

\section{Hypernode Classification Experiments} \label{sec:experiments}

In this section we evaluate $\text{PhenomNN}_{\text{simple}}$ and PhenomNN on various hypergraph benchmarks focusing on hypernode classification and compare against previous SOTA approaches. 
% \subsection{Hypernode Classification on Benchmark Datasets}

\textbf{Datasets.} Existing hypergraph benchmarks mainly focus on hypernode classification. We adopt five public citation network datasets from \cite{Zhang2022HypergraphCN}: Co-authorship/Cora,Co-authorship/DBLP, Co-citaion/Cora, Co-citaion/Pubmed, Co-citaion/Citeseer. These datasets and splits are constructed by \cite{hypergcn} (\url{https://github.com/malllabiisc/HyperGCN}). We also adopt two other public visual object classification datasets: Princeton ModelNet40~\cite{wu20153d} and the National Taiwan University~(NTU) 3D model dataset~\citep{chen2003visual}. We follow HGNN~\citep{hgnn} to preprocess the data by MVCNN \citep{su2015multi} and GVCNN~\citep{feng2018gvcnn} and obtain the hypergraphs.  Additionally, we use the datasets provided by the public code~{(\url{https://github.com/iMoonLab/HGNN})} associated with \citep{hgnn}. 
Finally, \cite{chien2022you} construct a public hypergraph benchmark for hypernode classification which includes $\text{ModelNet40}^*$, $\text{NTU2012}^*$, Yelp~\citep{YelpDataset}, House~\citep{chodrow2021hypergraph}, Walmart~\citep{amburg2020hypergraph}, and 20News~\cite{dua2017uci}. $\text{ModelNet40}^*$ and $\text{NTU2012}^*$ have the same raw data as ModelNet40 and NTU2012 mentioned before in \cite{Zhang2022HypergraphCN} but different splits. All datasets from \cite{chien2022you} are downloaded from their code site (\url{https://github.com/jianhao2016/AllSet}).\footnote{Note that we excluded a few datasets for the following reasons: The Zoo dataset is very small; the Mushroom dataset is too easy; the Citation datasets are similar to \cite{Zhang2022HypergraphCN}, and since we have ModelNet40* and NTU2012* for comparison of different baselines from both papers, we did not select them.}

\textbf{Baselines.} For datasets from ~\cite{Zhang2022HypergraphCN}, we adopt the baselines from their paper which includes a multi-layer perceptron with explicit hypergraph Laplacian regularization (MLP+HLR), FastHyperGCN~\cite{hypergcn}, HyperGCN~\cite{hypergcn}, HGNN~\cite{hgnn}, HNHN~\cite{dong2020hnhn}, HGAT~\cite{HyperGAT}, HyperSAGE~\cite{arya2020hypersage}, UniGNN~\citep{UniGNN}, and various hypergraph GNNs (H-GNNs)~\cite{Zhang2022HypergraphCN} proposed by them. For datasets from ~\cite{chien2022you}, we also select baselines from their paper including an MLP, CE (Clique Expansion)$+$GCN, CE$+$GAT, HNHN, HGNN, HCHA~\cite{bai2021hypergraph}, HyperGCN, UniGCNII~\citep{UniGNN}, HAN~\cite{wang2019heterogeneous} with full batch and mini-batch settings, and AllsetTransformer and AllDeepSets~\cite{chien2022you}.

\textbf{Implementations.} We use a one-layer MLP for $f(X;W)$. Also, in practice we found that only using ReLU at the end of propagation steps works well. Detailed hyperparameter settings are deferred to Appendix \ref{appendix:hyperparams-phenomNN}. We choose the hidden dimension of our models to be the same or less than the baselines in previous work. For results in Table \ref{tab:icmlworkshopdataset}, we conduct experiments on 10 different train-test splits and report average accuracy of test samples following \cite{Zhang2022HypergraphCN}. For results in Table \ref{tab:iclr22dataset}, we randomly split the data into training/validation/test samples using (50\%/25\%/25\%) splitting percentages as in \cite{chien2022you} and report the average accuracy over ten random splits. All experiments are implemented on RTX 3090 with Pytorch and DGL~\cite{wang2019dgl}.

\textbf{Results.} As shown in Table \ref{tab:icmlworkshopdataset}, our models achieve the best performance and top ranking on all datasets from \cite{Zhang2022HypergraphCN} compared to previous baselines. And in Table \ref{tab:iclr22dataset}, our models achieve the first ($\text{PhenomNN}_{\text{simple}}$) and tied-for-second (PhenomNN) overall performance ranking on the benchmarks from \cite{chien2022you}. 

\textbf{Empirical evaluation of time and space complexity.} In practice, we find that PhenomNN is roughly 2$\times$ to 3$\times$ slower than a GCN given the integration of two expansions based on $H_0$ and $H_1$, which implies that the constant multiplying the theoretical complexity from above is at least doubled as expected.  Of course timing results will still vary based on hardware and implementation details. As an example, we measure the training time of GCN and our models on the same hardware on Coauthorship-DBLP data with hidden size 64 and 8 layers.  We observe 0.047s/epoch for GCN and 0.045s/epoch for PhenomNN$_\text{simple}$ and 0.143s/epoch for PhenomNN under these conditions.  In terms of the space efficiency, our models are also analytically similar to common GNNs. And under the same settings as above, the memory consumption is 1665MB for GCN, 1895MB for PhenomNN$_\text{simple}$, and 2424MB for PhenomNN.

\textbf{Ablations.} For space considerations, we defer ablations to Appendix \ref{appendix:ablations}; however, we nonetheless highlight some of our findings here. For example, in Table \ref{tab:ablation1} (Appendix \ref{appendix:ablations}) we demonstrate the effect of different hypergraph energy function terms, which are associated with different hypergraph expansions per Proposition \ref{proposition:energy}.  In brief here, we explore different selections of $\{\lambda_0,\lambda_1\} \in \{\{0,1 \},\{1,0 \},\{1,1 \}, \}$ which in effect modulate the inclusion of clique- and star-like expansion factors.  Results demonstrate that on most datasets, the combination of both expansions, with their complementary roles, is beneficial. 

We also explore the tolerance of our model to different hidden dimensions in Table \ref{tab:hidden}. In brief, we fix other hyperparameters and obtain results across different hidden dimensions with PhenomNN$_\text{simple}$ for simplicity; results for PhenomNN are similar.  Overall, this ablation demonstrates the stability of our approach across hidden dimension.

\textbf{Additional comparisons and discussion.} As suggested by reviewers, we include additional discussion and comparison with existing work in Appendix \ref{appendix:newresults} due to the page limit.  This includes side-by-side evaluations with RGCN and the model from \cite{wang2022equivariant} which was not yet published at the time of our original submission.

\section{Conclusion}
While hypergraphs introduce compelling modeling flexibility, they still remain relatively under-explored in the GNN literature. With the potential to better understand hypergraph properties and expand their utility, we have introduced an expressive family of hypergraph energy functions and fleshed out their connection with previous hypergraph expansions.  We then leverage this perspective to design what can be interpreted as hypergraph neural network layers that are in one-to-one correspondence with  proximal gradient steps descending these energies.  We also characterize the similarities and differences of these layers w.r.t.~popular existing GNN architectures.  In the end, the proposed framework achieves competitive or SOTA performance on key hypergraph node classification benchmarks.  

\section{Acknowledgments}
This work was supported by the National Natural Science Foundation of China (No. 62236004 and No. 62022027) and the Major Key Project of PCL (PCL2021A12).

\bibliography{references,wipf_refs}
\bibliographystyle{icml2023}

%%%%%%%%%%%%%%%%%%%%%%%%%%%%%%%%%%%%%%%%%%%%%%%%%%%%%%%%%%%%%%%%%%%%%%%%%%%%%%%
%%%%%%%%%%%%%%%%%%%%%%%%%%%%%%%%%%%%%%%%%%%%%%%%%%%%%%%%%%%%%%%%%%%%%%%%%%%%%%%
% APPENDIX
%%%%%%%%%%%%%%%%%%%%%%%%%%%%%%%%%%%%%%%%%%%%%%%%%%%%%%%%%%%%%%%%%%%%%%%%%%%%%%%
%%%%%%%%%%%%%%%%%%%%%%%%%%%%%%%%%%%%%%%%%%%%%%%%%%%%%%%%%%%%%%%%%%%%%%%%%%%%%%%
\newpage
\appendix
\onecolumn

\appendix

\section{Additional Comparisons with Existing Work}
\label{appendix:newresults}
\subsection{Further Discussion}
In terms of expressiveness and generalizability, the primary difference between the AllSet from~\cite{chien2022you} and PhenomNN can be loosely distilled as follows: AllSet is explicitly designed for expanding per-layer expressive power as much as possible by combining principles from Deep Sets and SetTransformers. But this complexity may reduce the feasibility of exploring deeper models that spread information longer distances across a hypergraph, e.g., only a single layer model is used for the experiments in~\cite{chien2022you}. In contrast, PhenomNN is motivated by harnessing the interpretable inductive biases that come from descending an explicit lower-level energy function, whose minima are trainable by a higher-level downstream classification task. In this way, PhenomNN can in principle include an arbitrary number of layers to pass information across the hypergraph, since additional layers merely iterate the embeddings closer to the energy function minimum. Given these considerations, both AllSet and PhenomNN both have merits, and neither model fully encompasses the other as a special case.

Another recently published work raised by reviewers \cite{wang2022equivariant} proposes a quite interesting model called ED-HNN (for equivariant diffusion hypergraph neural network). This approach is complementary to our submission and different in at least three key respects:  First, although \cite{wang2022equivariant} use gradient diffusion processes to motivate a broad class of GNN models, that are in some ways similar to AllSets from \cite{chien2022you}, in the end there is not actually any specific energy function that is being minimized by their proposed Algorithm 1. Indeed there is no guarantee provided that each layer of their method is reducing any specific graph-regularized quantity of interest, which is our primary focus.

Secondly, their incorporation of proximal operators is fundamentally different than ours. In ED-HNN,  MLPs are used to implicitly model the effect of arbitrary proximal operators within an iterative ADMM optimization scheme. However, although conceptually understandable, a general MLP can model any function while proximal operators must obey very stringent properties (e.g., in 1D they must be nondecreasing functions of the input argument). In contrast, we apply proximal gradient descent to an explicit energy function with strict convergence guarantees.  And last but not least, we consider an energy function dependent on both node and hyperedge embeddings, while ED-HNN only considers node-wise embeddings (that are regrouped within a penalty for each hyperedge).

\subsection{Extra Empirical Results}
\textbf{ED-HNN comparisons.}  We include five new benchmarks for comparison from ED-HNN~\cite{wang2022equivariant} suggested by reviewers in Table \ref{tab:ablation-moredatasets} , where we observe that both models perform well relative to a wide variety of baselines.
\begin{table*}[htb]
\def\p{ ± } 
\centering
\caption{Extension datasets of Table \ref{tab:iclr22dataset} to be compared with ED-HNN. Results for ED-HNN are from ~\cite{wang2022equivariant}, while results for other baselines are from ~\cite{chien2022you}.}
\setlength\tabcolsep{2.5pt}

    \begin{tabular}{l|c|c|c|c|c}
        \toprule 

            \hline
            \hline
             &
  Cora &
  Citeseer &
  Pubmed &
  Cora-CA &
  DBLP-CA \\
  \hline
  MLP &
  75.17  ±  1.21 &
 72.67  ±  1.56 &
  87.47  ±  0.51 &
  74.31  ±  1.89 &
  84.83  ±  0.22 
  \\
CECGN &
  76.17  ±  1.39 &
  70.16  ±  1.31 &
  86.45  ±  0.43 &
  77.05  ±  1.26 &
  88.00  ±  0.26  \\
CEGAT &
  76.41  ±  1.53 &
  70.63  ±  1.30 &
  86.81  ±  0.42 &
  76.16  ±  1.19 &
  88.59  ±  0.29  \\
HNHN &
  76.36  ±  1.92 &
  72.64  ±  1.57 &
  86.90  ±  0.30 &
  77.19  ±  1.49 &
  86.78  ±  0.29  \\
HGNN &
  79.39  ±  1.36 &
  72.45  ±  1.16 &
  86.44  ±  0.44 &
  82.64  ±  1.65 &
  91.03  ±  0.20 
  \\
HCHA &
 79.14  ±  1.02 &
  72.42  ±  1.42 &
  86.41  ±  0.36 &
  82.55  ±  0.97 &
  90.92  ±  0.22  \\
HyperGCN &
  78.45  ±  1.26 &
  71.28  ±  0.82 &
  82.84  ±  8.67 &
  79.48  ±  2.08 &
  89.38  ±  0.25  \\
UniGCNII &
  78.81  ±  1.05 &
  73.05  ±  2.21 &
  88.25  ±  0.40 &
  83.60  ±  1.14 &
 91.69  ±  0.19  \\
HAN (full batch)$^*$ &
 80.18  ±  1.15&
  74.05  ±  1.43 &
  86.21  ±  0.48 &
 84.04  ±  1.02 &
  90.89  ±  0.23  \\
HAN (mini batch)$^*$ &
  79.70  ±  1.77 &
  74.12  ±  1.52 &
  85.32  ±  2.25 &
  81.71  ±  1.73 &
  90.17  ±  0.65 \\
AllDeepSets  & 76.88  ±  1.80 & 70.83  ±  1.63 & \underline{88.75  ±  0.33} & 81.97  ±  1.50 & 91.27  ±  0.27 \\
AllSetTransformer  & 78.58  ±  1.47 & 73.08  ±  1.20 & 88.72  ±  0.37 & 83.63  ±  1.47 & 91.53  ±  0.23  \\
\midrule
ED-HNN & 80.31  ±  1.35 & 73.70  ±  1.38& \textbf{89.03  ±  0.53} & 83.97  ±  1.55 & 91.90  ±  0.19  \\

   $\text{PhenomNN}_{\text{simple}}$&
    \underline{81.98  ±  1.58	}&\underline{75.00  ±  0.58}	&88.25  ±  0.42	&\underline{85.18  ±  0.97}&	\underline{91.91  ±  0.24}
   \\
     PhenomNN &
        \textbf{82.29  ±  1.42}&	\textbf{75.10  ±  1.59}&	88.07  ±  0.48&	\textbf{85.81  ±  0.90}&	\textbf{91.91  ±  0.21}

   \\
        \bottomrule

    \end{tabular}
    
\label{tab:ablation-moredatasets} 
\end{table*}

\textbf{RGCN comparisons.} Although we already provide comparisons with the heterogeneous GNN model HAN in Table \ref{tab:iclr22dataset}, given the connection between PhenomNN and RGCN models detailed in Section \ref{sec:rgcn}, it also makes sense to provide further evaluations with the latter.  To this end, Table \ref{tab:RGCN} summarizes results comparing PhenomNN to heterogeneous graphs applied to the star expansion. HAN results are reproduced from the main paper.  Meanwhile, for the RGCN for hypergraphs implementation, we use code from ~\cite{chien2022you} which executes full-batch training that produces OOM for some datasets. In any event, from the available results we observe that our models can outperform both RGCN and the related heterogeneous GNN HAN models alike.
\begin{table*}[htbp]
\centering
\setlength{\tabcolsep}{2.5pt}
\caption{Additional comparisons with heterogeneous GNN models applied to star expansions.} % % OOM indicates out-of-memory. For computing rank, OOM is put in the last place.
\label{tab:RGCN} 
\scriptsize
\begin{tabular}{@{}c|cccccccc@{}}
\toprule

\multicolumn{1}{l}{} &
  $\text{NTU2012}^{*}$ &
  $\text{ModelNet40}^{*}$ &
  Yelp &
  House(1) &
  Walmart(1) &
  House(0.6) &
  Walmart(0.6) &
  20Newsgroups
   \\ \midrule

HAN (full batch)$^*$ &
  83.58  ±  1.46 &
  94.04  ±  0.41 &
  OOM &
  \underline{71.05  ±  2.26} &
  OOM &
  83.27  ±  1.62 &
  OOM & OOM
   \\
HAN (mini batch)$^*$ &
  80.77  ±  2.36 &
  91.52  ±  0.96 &
  26.05  ±  1.37 &
  62.00  ±  9.06 &
  48.57  ±  1.04 &
  82.04  ±  2.68 &
  63.10  ±  0.96 & 79.72  ±  0.62
  \\
RGCN (full batch)$^*$ &
  86.74  ±  1.69&	97.62  ±  0.32&	OOM	&66.38  ±  3.69&	OOM&	78.17  ±  2.74&	OOM&	OOM
   \\

  \hline
      $\text{PhenomNN}_{\text{simple}}$ &
  \textbf{91.03  ±  1.04} &
  \textbf{98.66  ±  0.20} &
\textbf{32.26  ±  0.40}&
   \textbf{71.77  ±  1.68} &
  \textbf{64.11  ±  0.49} &
 \textbf{86.96  ±  1.33} &
 \textbf{78.46  ±  0.32} &
\textbf{81.74  ±  0.52} 
   \\
      PhenomNN &
  \underline{90.62  ±  1.88} &
  \underline{98.61  ±  0.17} &
\underline{31.92  ±  0.36} &
  70.71  ±  2.35  &
  \underline{62.98  ±  1.36}  &
  \underline{85.28  ±  2.30} &
 \underline{78.26  ±  0.26} &
\underline{81.41  ±  0.49} 
\\

  \bottomrule
  
\end{tabular}

\end{table*}

\section{Ablation Tables}
\label{appendix:ablations}
% We put the results table for combination ablation here.
\begin{table*}[htb]
\def\p{ ± } 
\centering
\caption{Results for ablations of hypergraph expansion combinations under the same settings as applied in the main paper. Boldfaced letters indicate the best expansion compared with the same model.}
\setlength\tabcolsep{2.5pt}

\scriptsize
    \begin{tabular}{l|c|c|c|c|c|c|c|c}
        \toprule 
              \multicolumn{1}{c}{}&\multicolumn{1}{c}{\begin{tabular}[c]{@{}c@{}}Cora\\ (co-authorship)\end{tabular}}  & \multicolumn{1}{c}{\begin{tabular}[c]{@{}c@{}}DBLP\\ (co-authorship)\end{tabular}} & \multicolumn{1}{c}{\begin{tabular}[c]{@{}c@{}}Cora\\ (co-citation)\end{tabular}} & \multicolumn{1}{c}{\begin{tabular}[c]{@{}c@{}}Pubmed\\ (co-citation)\end{tabular}} & \multicolumn{1}{c}{\begin{tabular}[c]{@{}c@{}}Citeseer\\ (co-citation)\end{tabular}}& \multicolumn{1}{c}{\begin{tabular}[c]{@{}c@{}}NTU2012\\ (both features)\end{tabular}}&\multicolumn{1}{c}{\begin{tabular}[c]{@{}c@{}}ModelNet40\\ (both features)\end{tabular}}  \\
        \midrule
        $\text{PhenomNN}_{\text{simple}}$-\textbf{clique}  & 77.06  ±  1.27 & 89.54  ±  0.05 & 72.37  ±  1.49 & 75.71  ±  1.04& 64.92  ±  1.56 & 85.36 ±  0.36 & 97.81 ± 0.09 \\
        $\text{PhenomNN}_{\text{simple}}$-\textbf{star}  & 77.28  ±  1.27 & 89.54  ±  0.18 & \textbf{72.81  ±  1.67} & \textbf{76.20  ±  1.41} & 64.96  ±  1.13 & 85.31 ±  0.23  & 97.81 ± 0.08  \\
    $\text{PhenomNN}_{\text{simple}}$  & \textbf{77.62  ±  1.30} & \textbf{89.74  ±  0.16} &\textbf{72.81  ±  1.67}
            & \textbf{76.20  ±  1.41} & \textbf{65.07  ±  1.08} & \textbf{85.39 ± 0.40}& \textbf{97.83 ± 0.09}\\
            \hline
    PhenomNN-\textbf{clique}  & 76.74  ±  0.41 & 89.56  ±  0.08 &72.68  ±  0.63
            & 77.94  ±  0.20 & 65.65  ±  0.34 &  85.15 ±  0.40 &97.71 ± 0.15\\
PhenomNN-\textbf{star}  & 76.83  ±  0.52 & 89.52  ±  0.05 &\textbf{73.09  ±  0.65}
            & 77.52  ±  0.34 & 65.46  ±  0.46 & 85.25 ±  0.38  & \textbf{97.77 ± 0.11}\\
PhenomNN  & \textbf{77.11  ±  0.45} & \textbf{89.81  ±  0.05} &\textbf{73.09  ±  0.65}
            & \textbf{78.12  ±  0.24} & \textbf{65.77  ±  0.45}& \textbf{85.40 ± 0.42} & \textbf{97.77 ± 0.11}\\

            \hline
            \hline
            \multicolumn{1}{c}{} &
  NTU2012* &
  ModelNet40* &
  Yelp &
  House(1) &
  Walmart(1) &
  House(0.6) &
  Walmart(0.6) &
  20Newsgroups \\
  \hline
$\text{PhenomNN}_{\text{simple}}$-\textbf{clique} &
  90.36  ±  1.80 &
  98.64  ±  0.23 &
31.76  ±  0.42&
   70.93  ±  2.25 &
   61.84  ±  0.66 &
 86.40  ±  1.60 &
 77.38  ±  0.17 &
\textbf{81.74  ±  0.52}
   \\

   $\text{PhenomNN}_{\text{simple}}$-\textbf{star} &
  90.68  ±  1.38 &
  98.50  ±  0.13 &
 32.18  ±  0.41&
  71.21  ±  2.19 &
  \textbf{64.11  ±  0.49} &
 86.26  ±  1.51 &
 78.38  ±  0.21 &
81.47  ±  0.38
   \\
      $\text{PhenomNN}_{\text{simple}}$ &
  \textbf{91.03  ±  1.04} &
  \textbf{98.66  ±  0.20} &
\textbf{ 32.26  ±  0.40}&
   \textbf{71.77  ±  1.68} &
  \textbf{64.11  ±  0.49} &
 \textbf{86.96  ±  1.33} &
 \textbf{78.46  ±  0.32} &
\textbf{81.74  ±  0.52}

   \\\hline
   PhenomNN-\textbf{clique} &
  90.14  ±  1.26 &
 98.55  ±  0.16 &
31.58  ±  0.53 &
   70.37  ±  2.66  &
 60.96  ±  0.37  &
 85.00 ±  1.82 &
77.19  ±  0.25 &
81.07  ±  0.54
   \\   
   PhenomNN-\textbf{star} &
  90.38  ±  1.78 &
 \textbf{98.61  ±  0.17} &
\textbf{31.92  ±  0.36} &
   69.50  ±  2.34  &
 \textbf{63.82 ±  0.49}  &
 \textbf{85.22  ±  1.67} &
\textbf{78.26  ±  0.26} &
81.11  ±  0.36
   \\   
   PhenomNN &
  \textbf{90.62  ±  1.88} &
  \textbf{98.61  ±  0.17} &
\textbf{31.92  ±  0.36}&
  \textbf{70.71  ±  2.35}  &
  \textbf{ 63.82 ±  0.49 } &
  \textbf{85.22  ±  1.67}&
 \textbf{78.26  ±  0.26} &
\textbf{81.41  ±  0.49}
   \\
        \bottomrule

    \end{tabular}
    
\label{tab:ablation1} 
\end{table*}

\begin{table*}[htb]
\def\p{ ± } 
\centering
\caption{Results with different hidden sizes of $\text{PhenomNN}_{\text{simple}}$. NTU2012*, ModelNet40*, House(1), and House(0.6) are four representative datasets from Table \ref{tab:iclr22dataset} in the main paper. The '/' symbol indicates that the result was not computed because this dimension is higher than what is used in our paper and previous works. }
\setlength\tabcolsep{2.5pt}

    \begin{tabular}{l|c|c|c|c|c}
        \toprule 

            \hline
            \hline
            Model& Hidden &
  NTU2012* &
  ModelNet40* &
  House(1) &
  House(0.6) \\
  \hline
  \multirow{4}*{$\text{PhenomNN}_{\text{simple}}$
                        }
&512 &
    /	&98.66  ± 0.20&	71.77  ±  1.68&	86.96  ±  1.33
   \\

   ~&256&
   91.03  ±  1.04&	98.57  ±  0.14&	69.38  ±  2.47&	86.12  ±  2.11

   \\
     ~&128 &
    89.46  ±  1.39&	98.42  ±  0.15&	68.60  ±  1.96	&84.56  ±  1.42

   \\
        ~&64 &
    89.96  ±  1.26&	98.51  ±  0.21&	68.66  ±  2.10&	85.44  ±  1.46

   \\
        \bottomrule

    \end{tabular}
    
\label{tab:hidden} 
\end{table*}
\newpage
\section{Proofs}
\label{appendix:proofs}
\subsection{Proof of Proposition \ref{proposition:energy}}
We first reproduce the proposition here for ease of comparison. 
Suppose  $g_2(Z,U;\psi)$ is removed from (\ref{eq:generalYU}), $H_0 = H_1 = I$, and define $Z^* \triangleq D_H^{-T} B^T Y$.  It then follows that
\begin{eqnarray}
& \hspace*{-1.0cm} \min_Z~\ell(Y,Z; \psi) ~~ = ~~ g_1(Y,X;\psi) + \sum_{i=1}^n \phi ( y_{i}) 
% &+ & \hspace*{-0.3cm} \lambda_0 \sum_{e_k \in \mathcal{E}}\sum_{i \in e_k}\sum_{j \in e_k}||y_i -y_j||^2_2 + \lambda_1 \sum_{e_k \in \mathcal{E}}\sum_{i \in e_k}||y_i-z^*_k||^2_2 \nonumber \\
% & = &   g_1(Y,X;\psi) + \sum_{i=1}^n \phi ( y_{i}) \nonumber \\
+   2\lambda_0 \mbox{tr}[Y^T L_{C} Y] + \lambda_1 \text{tr}\left(\left[\begin{array}{c}
         Y  \\
         Z^* 
    \end{array}\right]^TL_{S}\left[\begin{array}{c}
         Y  \\
         Z^*
    \end{array}\right] \right) \label{eq:firsteqmin}\\
    &  \hspace*{-1.0cm} =  g_1(Y,X;\psi) + \sum_{i=1}^n \phi ( y_{i}) + 2\lambda_0 \mbox{tr}[Y^T L_{C} Y] + \lambda_1 \mbox{tr}[Y^T \bar{L}_S Y]\label{eq:secondeqmin}
\end{eqnarray}
    % & + & \lambda_0 \mbox{tr}[Y^T L_{C} Y] + \lambda_1 \mbox{tr}[Y^T \bar{L}_{S} Y]
    Moreover, if $\calG$ is $m_e$-uniform, then under the same assumptions
\begin{equation}
\label{eq:moreresults}
 \min_Z~\ell(Y,Z; \psi)  = g_1(Y,X;\psi) + \sum_{i=1}^n \phi ( y_{i}) + \beta \mbox{tr}[Y^T L_{C} Y],
\end{equation}
where $\beta \triangleq 2\lambda_0 + \tfrac{\lambda_1}{m_e}$.
\begin{proof}
After removing $g_2(Z,U;\psi)$ from $\ell(Y,Z;\psi)$ and setting $H_0=H_1=I$, we have
\begin{align}
\label{eq:energywithoutg2}
   \ell(Y,Z;\psi)=g_1(Y,X;\psi)+ \sum_{i=1}^n \phi ( y_{i})+\sum_{k=1}^m \phi ( z_{i}) + 
    \lambda_0\sum_{e_k \in \mathcal{E}}\sum_{i \in e_k}\sum_{j \in e_k}||y_i-y_j||^2_2  
        +\lambda_1\sum_{e_k \in \mathcal{E}}\sum_{i \in e_k}||y_i-z_k||^2_2
\end{align}
First we know 
\begin{equation}
    \sum_{e_k \in \mathcal{E}}\sum_{i \in e_k}\sum_{j \in e_k}||y_i-y_j||^2_2=2\mbox{tr}[Y^T L_{C} Y]
\end{equation} from the definition of Laplacian $L_C$. Then we solve $Z^*$ for minimizing \eqref{eq:energywithoutg2}. If there were no $\phi$ term, then it follows that $z^*_k=\text{MEAN}({y_i| i\in e_k})$, because the mean function minimizes the sum of squared errors.  However, because each $y_i$ is forced to be positive by $\phi$, the resulting mean will also be positive and therefore feasible as well.  Therefore, the mean estimator will remain optimal even if we include the $\phi$ term.  

It can also be shown that the aforementioned mean estimator satisfies 
\begin{equation}
    Z^* = D_H^{-T} B^T Y,
\end{equation}
and hence
\begin{equation}
    \sum_{e_k \in \mathcal{E}}\sum_{i \in e_k}||y_i-z^*_k||^2_2 = \text{tr}\left(\left[\begin{array}{c}
         Y  \\
         Z^* 
    \end{array}\right]^TL_{S}\left[\begin{array}{c}
         Y  \\
         Z^*
    \end{array}\right] \right) 
\end{equation}
from the definition of Laplacian $L_S$.  This expression then allows us to reproduce \eqref{eq:firsteqmin}.
% And since $z^*_k$ is the mean of several non-negative node embeddings regularized by $\phi$, $z^*_k$ is also non-negative. Thus we can eliminate $\phi(z_i)$ without loss of generality.
Processing further, we have
\begin{align}
    \sum_{e_k \in \mathcal{E}}\sum_{i \in e_k}||y_i-z^*_k||^2_2&=\sum_{e_k \in \mathcal{E}}\sum_{i \in e_k}||y_i-\frac{\sum_{j\in e_k}y_j}{m_{e_k}}||^2_2 \nonumber\\
    &=\sum_{e_k \in \mathcal{E}}\frac{1}{m_{e_k}}\sum_{i \in e_k}\sum_{j\in e_k}||y_i-y_j||^2_2 \nonumber\\
    &=\text{tr}[Y^T\bar{L}_SY],
\end{align}
which leads to \eqref{eq:secondeqmin}.

Recall the definition of $A_C=BB^T$ and $\bar{A}_S=BD_H^{-1}B^T$, so if $\mcG$ is $m_e$-uniform, it means all diagonal elements in $D_H$ is $m_e$, so we have
\begin{align}
    \bar{A}_S=BD_H^{-1}B^T=\frac{1}{m_e}BB^T=\frac{1}{m_e}A_C.
\end{align}
From the definition of $\bar{L}_S$, we get 
\begin{align}
    \bar{L}_S&=\bar{D}_S -\bar{A}_S\nonumber\\
    &=\frac{1}{m_e}D_C - \frac{1}{m_e}A_C \nonumber\\
    &=\frac{1}{m_e}L_C,
\end{align}
which leads to \eqref{eq:moreresults}.
\end{proof}

\subsection{Proof of Proposition \ref{proposition:gradientconverge}}
It is notable that the updating consists of two parts \eqref{eq:before_prox} and \eqref{eq:after_prox} using the proximal gradient descent we discussed before. So the main point here is to prove the descent of \eqref{eq:before_prox} for $\psi=\{W,H_0,H_1\}$ for Proposition \ref{proposition:gradientconverge} and $\psi=\{W,I,I\}$ for Corollary \ref{corollary:gradientconverge-simple} respectively.

We first provide a basic mathematical result.

\begin{lemma}\label{lemma:RothColumn-supp} (Roth's Column Lemma \cite{(RothColumn)Harold1981}). For any three matrices $\mbX, \mbY$ and $\mbZ$,
\begin{align}
    vec(\mbX\mbY\mbZ) = (\mbZ^{\top}\otimes\mbX) vec(\mbY) \label{eq:RothColumn-supp}
\end{align}
\end{lemma}

We now proceed with the proof of our result.

\begin{proof}
The gradient of $\bar{\ell}(Y;\psi)$ is as follows:
\begin{align}
\label{eq:gradient}
    \nabla_{Y}\bar{\ell}(Y;\psi) = 2\left(\lambda_0(D_{C}Y-\Tilde{Y}_{C})+ \lambda_1(\bar{A}_{S}Y-\Tilde{Y}_{S}) + Y - f\left(X ; W  \right)\right),
\end{align}
where $\Tilde{Y}_{C}=A_{C}Y(H_0+H_0^T)-D_{C}YH_0H_0^T$ and $\Tilde{Y}_{S}=\bar{A}_{S}Y(H_0+H_0^T)-\bar{D}_{S}YH_0H_0^T$. We rewrite the equation in 
\begin{align}
\label{eq:gradient-supp}
   \frac{\nabla_{Y}\bar{\ell}(Y;\psi)}{2}= (\mbI + \lambda_0D_{C} + \lambda_1\bar{A}_{S})Y - f\left(X ; W  \right)  \nonumber\\ - 
    \lambda_0(A_{C}Y(H_0+H_0^T)-D_{C}YH_0H_0^T) \nonumber\\ - \lambda_1(\bar{A}_{S}Y(H_0+H_0^T)-\bar{D}_{S}YH_0H_0^T) ,
\end{align}
We do vectorization on both sides of (\ref{eq:gradient-supp}) to obtain:

\begin{align}
\label{eq:vectorization}
   \frac{vec(\nabla_{Y}\bar{\ell}(Y;\psi))}{2}= (\mbI +  \lambda_0I\otimes D_{C} + \lambda_1I\otimes \bar{A}_{S})vec(Y) - vec(f\left(X ; W  \right))  \nonumber\\ - vec\left(
    \lambda_0(A_{C}Y(H_0+H_0^T)-D_{C}YH_0H_0^T)\right) \nonumber\\ - vec\left(\lambda_1(\bar{A}_{S}Y(H_0+H_0^T)-\bar{D}_{S}YH_0H_0^T)\right) ,
\end{align}
Here, using Roth's Column Lemma \ref{lemma:RothColumn-supp} to rewrite equation \eqref{eq:vectorization}

\begin{align}
\label{eq:vectorization2}
   \frac{vec(\nabla_{Y}\bar{\ell}(Y;\psi))}{2}= (\mbI +  \lambda_0I\otimes D_{C} + \lambda_1I\otimes \bar{A}_{S})vec(Y) - vec(f\left(X ; W  \right))  \nonumber\\ - vec\left( 
    \lambda_0(H_0+H_0^T)\otimes A_{C}Y\right)+ vec\left(\lambda_0H_0^TH_0\otimes D_{C}Y\right) \nonumber\\ - vec\left(
    \lambda_1(H_0+H_0^T)\otimes \bar{A}_{S}Y\right)+ vec\left(\lambda_1H_0^TH_0\otimes \bar{D}_{S}Y\right),
\end{align}

This is needed behind but for now we just leave it. Then we write the updating after pre-conditioning in 
\begin{align} 
\label{eq:PhenomNN-H-updating-append}
\bar{Y}^{(t+1)} = Y^{(t)} - \alpha\tilde{D} ^{-1} \nabla_{Y^{(t)}}\bar{\ell}(Y;\psi),
\end{align}
where $\tilde{D}=\lambda_0 D_{C}+ \lambda_1 \bar{D}_{S} + I$.

Do vetorization on both sides turns \eqref{eq:PhenomNN-H-updating-append} to :
\begin{align}
    vec(\bar{Y}^{(t+1)}) = vec(Y^{(t)}) - \alpha vec(\tilde{D} ^{-1} \nabla_{Y^{(t)}}\bar{\ell}(Y;\psi)) \label{eq:vec_unfolding_precond-supp} \\
    =vec(Y^{(t)}) - \alpha \hat{D}^{-1} vec(\nabla_{Y^{(t)}}\bar{\ell}(Y;\psi)), \label{eq:vec_unfolding_precond_roth-supp}
\end{align}
where $\hat{D}^{-1}=I \otimes \tilde{D} ^{-1}$. Note that we apply Roth's column lemma to (\ref{eq:vec_unfolding_precond-supp}) to derive (\ref{eq:vec_unfolding_precond_roth-supp}).

From the property of strongly convex function $\ell (Y ; \psi)$, We know the following inequality holds for any $\bar{Y}^{(t+1)}$ and $Y^{(t)}$:
\begin{align}
    \ell (\bar{Y}^{(t+1)};\psi) \leq \ell (Y^{(t)};\psi) &+ vec(\nabla_{Y^{(t)}}\bar{\ell}(Y;\psi))^{\top} vec(\bar{Y}^{(t+1)} - Y^{(t)}) \nonumber\\
    &+ \frac{1}{2} vec(\bar{Y}^{(t+1)} - Y^{(t)})^{\top} \nabla_{Y^{(t)}}^2 \bar{\ell}(Y;\psi) vec(\bar{Y}^{(t+1)} - Y^{(t)}), \label{ineq:second_order}
\end{align}
where $\nabla^2_{Y^{(t)}} \bar{\ell}(Y;\psi)$ is a Hessian matrix whose elements are $\nabla^2_{Y^{(t)}} \bar{\ell}(Y;\psi)_{ij}=\frac{\partial \ell _{Y}(Y)}{\partial vec(Y)_i \partial vec(Y)_j} |_{Y=Y^{(t)}}$.

Applying the gradient descent update $vec(\bar{Y}^{(t+1)} - Y^{(t)}) = - \alpha \hat{D}^{-1} vec(\nabla_{Y^{(t)}}\bar{\ell}(Y;\psi)) $, we get:

\begin{align}
    \ell (\bar{Y}^{(t+1)};\psi) \leq \ell (Y^{(t)};\psi) &- ( \hat{D}^{-1} vec(\nabla_{Y^{(t)}}\bar{\ell}(Y;\psi)))^{\top} (\alpha \hat{D}) ( \hat{D}^{-1} vec(\nabla_{Y^{(t)}}\bar{\ell}(Y;\psi))) \nonumber\\
    &+( \hat{D}^{-1} vec(\nabla_{Y^{(t)}}\bar{\ell}(Y;\psi)))^{\top} (\frac{\alpha^2}{2}\nabla_{Y^{(t)}}^2 \bar{\ell}(Y;\psi))  ( \hat{D}^{-1} vec(\nabla_{Y^{(t)}}\bar{\ell}(Y;\psi))).
\end{align}

If $\alpha \hat{D} - \frac{\alpha^2}{2}\nabla^2_{Y^{(t)}} \bar{\ell}(Y;\psi) \succ 0$ holds, then gradient descent will always decrease the loss, and furthermore, since $\ell (Y ; \psi)$ is strongly convex, with proximal descent, it will monotonically decrease the loss until the unique global minimum.  To compute $\nabla^2_{Y^{(t)}}\bar{\ell}(Y;\psi)$, we differentiate (\ref{eq:vectorization2}) and arrive at:
\begin{align}
    \nabla^2_{Y^{(t)}}\bar{\ell}(Y;\psi) = 2(\mbI + Q-P+\mbD).
\end{align}
where $\mbD=  \lambda_0I\otimes D_{C} + \lambda_1I\otimes \bar{A}_{S} $ and $Q$ and $P$ is in Proposition \ref{proposition:gradientconverge}.
Returning to the above inequality, we can then proceed as follows:
\begin{align}
\label{eq:non-negative1}
    \alpha \hat{D} - \frac{\alpha^2}{2}(\mbI + Q-P+\mbD) &= \alpha (\mbI+\lambda_0I \otimes D_{C}+\lambda_1 I \otimes \bar{D}_{S}) - \alpha^2(\mbI + Q-P+\lambda_0I\otimes D_{C} + \lambda_1I\otimes \bar{A}_{S} ) \nonumber\\
    &= (\alpha - \alpha^2)(\mbI + \lambda_0 I \otimes D_{C}) + \lambda_1 I \otimes \alpha \bar{D}_{S} - \alpha^2 (Q - P+ \lambda_1I\otimes \bar{A}_{S}) \nonumber\\
    &\succ [(\alpha - \alpha^2)(1 + \lambda_0 d_{C\text{min}})+\alpha \lambda_1d_{S\text{min}}  ]\mbI - \alpha^2 (Q - P + \lambda_1I\otimes \bar{A}_{S}).
\end{align}

If $\alpha$ satisfies $[(\alpha - \alpha^2)(1 + \lambda_0 d_{C\text{min}})+\alpha \lambda_1d_{S\text{min}}  ]\mbI - \alpha^2 (Q - P + \lambda_1I\otimes \bar{A}_{S}) \succ 0$, then $\alpha \hat{D} - \frac{\alpha^2}{2}\nabla^2_{Y^{(t)}} \bar{\ell}(Y;\psi) \succ 0$ holds. Therefore, a sufficient condition for convergence to the unique global optimum is:
\begin{align}
    (\alpha - \alpha^2)(1 + \lambda_0 d_{C\text{min}})+\alpha \lambda_1d_{S\text{min}}   - \alpha^2 \sigma_{\text{max}} > 0.
\end{align}
where $\sigma_{\text{max}} \text{ is the max eigenvalue of }(Q - P + \lambda_1I\otimes \bar{A}_{S})$
Consequently, to guarantee the aforementioned convergence we arrive at the final inequality:
\begin{align}
    \alpha < \frac{1 + \lambda_0 d_{C\text{min}} + \lambda_1d_{S\text{min}}}{1 + \lambda_0 d_{C\text{min}} + \sigma_{\text{max}}}.
\end{align}
\end{proof}
\subsection{Proof for Corollary \ref{corollary:gradientconverge-simple}}
This proof is more simpler because without compatibility matrix we don't need vectorization here. For $\psi=\{W,I,I\}$, the gradient  becomes
\begin{align}
    \label{eq:gradient-simple}
    \nabla_{Y}\bar{\ell}(Y;\psi) = 2(\lambda_0L_{C}+\lambda_1\bar{L}_{S}) Y +
2Y - 2 f\left(X ; W  \right),
\end{align}
The Hessian matrix is 
\begin{align}
    \nabla^2_{Y^{(t)}}\bar{\ell}(Y;\psi) = 2(\lambda_0L_C+\lambda_1L_s+I).
\end{align}
While all other conditions are similar, we rewrite \eqref{eq:non-negative1} in 
\begin{align}
    \alpha \hat{D} - \frac{\alpha^2}{2}\nabla^2_{Y^{(t)}} \bar{\ell}(Y;\psi)=\alpha(\lambda_0D_{C}+\lambda_1\bar{D}_{S}+I) - \alpha^2(\lambda_0L_{C}+\lambda_1\bar{L}_{S}+I)\nonumber\\
    =(\alpha-\alpha^2)(\lambda_0D_{C}+\lambda_1\bar{D}_{S}+I) +\alpha^2(\lambda_0A_{C}+\lambda_1\bar{A}_{S}) \nonumber\\
    \succ (\alpha-\alpha^2)(\lambda_0d_{Cmin}+\lambda_1d_{Smin}+I) +\alpha^2(\lambda_0A_{C}+\lambda_1\bar{A}_{S})
\end{align}
To make $(\alpha-\alpha^2)(\lambda_0d_{Cmin}+\lambda_1d_{Smin}+I) +\alpha^2(\lambda_0A_{C}+\lambda_1\bar{A}_{S}) \succ 0$ a sufficient condition is that 
\begin{align}
    (\alpha-\alpha^2)(\lambda_0d_{Cmin}+\lambda_1d_{Smin}+1) +\alpha^2\sigma_{min} > 0
\end{align}
where $\sigma_{min}$ is the min eigenvalue of $(\lambda_0A_{C}+\lambda_1\bar{A}_{S})$.
That comes to 
\begin{align}
    \alpha < \frac{\lambda_0d_{Cmin}+\lambda_1d_{Smin}+1}{\lambda_0d_{Cmin}+\lambda_1d_{Smin}+1 -\sigma_{min}}.
\end{align}
\section{Hyperparameters }
Here we present hyperparameters for reproducing results in Table \ref{tab:icmlworkshopdataset}, Table \ref{tab:iclr22dataset} in Table \ref{tab:hyperparams-phenomNN} and \ref{tab:hyperparams-phenomNN-H} . And for Table \ref{tab:ablation1} the hyperparameters are in Table \ref{tab:hyperparams-phenomNN-ablation} and \ref{tab:hyperparams-phenomNN-H-ablation}. Note that in the ablation for combination coefficients, we re-searched for hyperparameters for each combination.
\label{appendix:hyperparams-phenomNN}
\begin{table}[htbp]
\centering
\scriptsize
\vspace{-0.3cm}
\setlength\tabcolsep{0.01\linewidth}
\renewcommand\arraystretch{1.1}

\caption{$\text{PhenomNN}_{\text{simple}}$ hyperparameters for Table \ref{tab:icmlworkshopdataset} and \ref{tab:iclr22dataset}.}
\begin{tabular}{cccccccc}
\toprule\toprule
Dataset & lr       & dropout & hidden & $\lambda_0$& $\lambda_1$ & $\alpha$ & prop step \\ \midrule\midrule
Coauthorship/Cora  & 0.01   & 0.7 & 64 &20 & 80 & 0.1 & 16\\
\hline
Coauthorship/DBLP  & 0.005   & 0.6 & 64 & 100& 100 & 0.1 & 16 \\
\hline
Cocitation/Cora  & 0.005   & 0.7 & 64 & 0& 20 & 1 & 16 \\
\hline
Cocitation/PubMed   & 0.02   & 0.7 & 64 &0 & 20 & 0.1 & 16 \\
\hline
Cocitation/Citeseer  & 0.005   & 0.7 & 64 &1 & 20 & 1 & 16 \\
\hline
NTU2012  & 0.001  & 0.2 & 128 &1 & 1 & 0.1 &16 \\
\hline
ModelNet40   & 0.0005  & 0.4 & 128 &1 & 1 & 0.05 &16 \\
\midrule\midrule

NTU2012*   & 0.01   & 0.2 & 256& 50 & 20 & 0.05 & 16 \\
\hline
ModelNet40*   & 0.01   & 0 & 512 &50 & 1 & 0.05 & 16  \\

\hline
Yelp  & 0.01   & 0.1 & 64 & 1& 100 & 0.1 & 4   \\
\hline
House(1)   & 0.1   & 0 & 512 &50 & 20 & $\frac{1}{70}$ or $(\lambda_0 + \lambda_1)^{-1}$ & 16   \\
\hline
House(0.6)   & 0.1   & 0 & 512&1 & 1 & 0.05 & 16 \\
\hline
Walmart(1)   & 0.01   & 0 & 256 & 0& 50 & 1 & 16 \\
\hline
Walmart(0.6)   & 0.1   & 0 & 256& 1 & 20 & 1 & 16  \\
\hline
20Newsgroups   & 0.01   & 0.2 & 64 &0.1 & 0 & 1 & 7   \\
\bottomrule\bottomrule
\end{tabular}\label{tab:hyperparams-phenomNN}
\end{table}
\begin{table}[htbp]
\centering
\scriptsize
\vspace{-0.3cm}
\setlength\tabcolsep{0.01\linewidth}
\renewcommand\arraystretch{1.1}

\caption{PhenomNN hyperparameters for Table \ref{tab:icmlworkshopdataset} and \ref{tab:iclr22dataset}.}
\begin{tabular}{cccccccc}
\toprule\toprule
Dataset & lr       & dropout & hidden & $\lambda_{0}$& $\lambda_{1}$ & $\alpha$ & prop step \\ \midrule\midrule
Coauthorship/Cora  & 0.001   & 0.8 & 64 &20 & 100 & 0.1 & 16\\

\hline
Coauthorship/DBLP  & 0.001   & 0.6 & 64 & 1& 1 & 1 & 16 \\

\hline
Cocitation/Cora  & 0.01   & 0.6 & 64 & 0& 20 & 1 & 16 \\

\hline
Cocitation/PubMed   & 0.01   & 0.6 & 64 &1 & 1 & 1 & 16 \\

\hline
Cocitation/Citeseer  & 0.001   & 0.8 & 64 &50 & 50 & 0.1 & 16 \\

\hline
NTU2012  & 0.001  & 0.2 & 64 &20 & 80 & 0.05 &16 \\

\hline
ModelNet40   & 0.0005  & 0.2 & 64 &0 & 20 & 0.05 &16 \\
\midrule\midrule

NTU2012*   & 0.01   & 0.2 & 256 & 100 & 20 & 0.05 & 16 \\

\hline
ModelNet40*   & 0.001   & 0.2 & 512 &0 & 20 & 0.05 & 16  \\

\hline
Yelp  & 0.01   & 0.2 & 64 & 0& 1 & 0.01 & 4   \\

\hline
House(1)   & 0.01   & 0.2 & 64 &50 & 100 & 0.05 & 16   \\

\hline
House(0.6)   & 0.01   & 0.2 & 512&0 & 1 & 0.05 & 16 \\

\hline
Walmart(1)   & 0.001   & 0 & 256 & 0& 50 & 1 & 16 \\

\hline
Walmart(0.6)   & 0.01   & 0 & 256& 0 & 50 & 1 & 16  \\

\hline
20Newsgroups   & 0.01   & 0 & 64 &0.1 & 0.1 & 1 & 8   \\

\bottomrule\bottomrule
\end{tabular}\label{tab:hyperparams-phenomNN-H}
\end{table}

\begin{table}[htbp]
\centering
\scriptsize
\vspace{-0.3cm}
\setlength\tabcolsep{0.01\linewidth}
\renewcommand\arraystretch{1.1}

\caption{$\text{PhenomNN}_{\text{simple}}$ hyperparameters for combination ablation. For every dataset, the first row is $\text{PhenomNN}_{\text{simple}}$-clique, the second is $\text{PhenomNN}_{\text{simple}}$-star.}
\begin{tabular}{cccccccc}
\toprule\toprule
Dataset & lr       & dropout & hidden & $\lambda_0$& $\lambda_1$ & $\alpha$ & prop step \\ \midrule\midrule
Coauthorship/Cora  & 0.01   & 0.7 & 64 &20 & 0 & 0.1 & 16\\
Coauthorship/Cora  & 0.01   & 0.7 & 64 &0 & 50 & 0.1 & 16\\
\hline
Coauthorship/DBLP  & 0.005   & 0.6 & 64 & 20 & 0 & 0.1 & 16 \\
Coauthorship/DBLP  & 0.01   & 0.6 & 64 & 0& 100 & 0.1 & 16 \\
\hline
Cocitation/Cora  & 0.005   & 0.6 & 64 & 20& 0 & 1 & 16 \\
Cocitation/Cora  & 0.005   & 0.7 & 64 & 0& 20 & 1 & 16 \\
\hline
Cocitation/PubMed  & 0.1   & 0.5 & 64 &20 & 0 & 1 & 16 \\
Cocitation/PubMed  & 0.02   & 0.7 & 64 &0 & 20 & 0.1 & 16 \\
\hline
Cocitation/Citeseer  & 0.01   & 0.7 & 64 &20 & 0 & 1 & 16 \\
Cocitation/Citeseer  & 0.005   & 0.7 & 64 &0 & 20 & 1 & 16 \\
\hline
NTU2012  & 0.001  & 0.2 & 128 &20 & 0 & 0.05 &16 \\
NTU2012   & 0.001  & 0.2 & 128 &0 & 100 & 0.1 &16 \\
\hline
ModelNet40    & 0.0005  & 0.4 & 128 &20 & 0 & 0.05 &16 \\
ModelNet40  & 0.0005  & 0.4 & 128 &0 & 20 & 0.05 &16 \\\midrule\midrule

NTU2012*  & 0.001   & 0 & 256 & 80 & 0 & 0.05 & 16 \\
NTU2012*  & 0.01   & 0 & 256& 0 & 1 & 0.1 & 16 \\
\hline
ModelNet40*   & 0.01   & 0.2 & 512 &100 & 0 & 0.05& 16  \\
ModelNet40*  & 0.01   & 0 & 512 &0 & 80 & 0.05 & 16  \\
\hline
Yelp   & 0.01   & 0 & 64 & 50& 0 & 0.01 & 4   \\
Yelp   & 0.01   & 0.1 & 64 & 0& 100 & 0.1 & 4   \\
\hline
House(1)   & 0.01   & 0 & 512 &80 & 0 & 0.05 & 16   \\
House(1)  & 0.01   & 0 & 512 &0 & 1 & 0.05 & 16   \\
\hline
House(0.6) & 0.1   & 0 & 512&1 & 0 & 0.05 & 16 \\
House(0.6)   & 0.1   & 0 & 512&0 & 80 & 0.05 & 16 \\
\hline
Walmart(1)   & 0.01   & 0 & 256 & 50& 0 & 1 & 16 \\
Walmart(1)  & 0.01   & 0 & 256 & 0& 50 & 1 & 16 \\
\hline
Walmart(0.6)   & 0.1   & 0 & 256& 20 & 0 & 1 & 16  \\
Walmart(0.6)   & 0.01   & 0 & 256& 0 & 20 & 1 & 16  \\
\hline
20Newsgroups   & 0.01   & 0.2 & 64 &0.1 & 0 & 1 & 7   \\
20Newsgroups   & 0.01   & 0.2 & 64 &0 & 0.1 & 1 & 7   \\
\bottomrule\bottomrule
\end{tabular}\label{tab:hyperparams-phenomNN-ablation}
\end{table}

\begin{table}[htbp]
\centering
\scriptsize
\vspace{-0.3cm}
\setlength\tabcolsep{0.01\linewidth}
\renewcommand\arraystretch{1.1}

\caption{PhenomNN hyperparameters for combination ablation. For every dataset, the first row is PhenomNN-clique, the second is PhenomNN-star.}
\begin{tabular}{cccccccc}
\toprule\toprule
Dataset & lr       & dropout & hidden & $\lambda_{0}$& $\lambda_{1}$ & $\alpha$ & prop step \\ \midrule\midrule
Coauthorship/Cora  & 0.001   & 0.8 & 64 &1 & 0 & 0.1 & 16\\
Coauthorship/Cora  & 0.001   & 0.8 & 64 &0 & 80 & 0.1 & 16\\
\hline
Coauthorship/DBLP  & 0.01   & 0.6 & 64 & 1 & 0 & 1 & 16 \\
Coauthorship/DBLP  & 0.01   & 0.6 & 64 & 0& 20 & 1 & 16 \\
\hline
Cocitation/Cora  & 0.01   & 0.6 & 64 & 50& 0 & 0.1 & 16 \\
Cocitation/Cora  & 0.01   & 0.6 & 64 & 0& 20 & 1 & 16 \\
\hline
Cocitation/PubMed  & 0.01   & 0.6 & 64 &1 & 0 & 1 & 16 \\
Cocitation/PubMed  & 0.001   & 0.8 & 64 &0 & 1 & 1 & 16 \\
\hline
Cocitation/Citeseer  & 0.001   & 0.8 & 64 &80 & 0 & 0.05 & 16 \\
Cocitation/Citeseer  & 0.001   & 0.8 & 64 &0 & 20 & 1 & 16 \\
\hline
NTU2012  & 0.001  & 0.2 & 64 &1 & 0 & 0.05 &16 \\
NTU2012   & 0.001  & 0.2 & 64 &0 & 100 & 0.1 &16 \\
\hline
ModelNet40    & 0.001  & 0.4 & 64 &1 & 0 & 0.05 &16 \\
ModelNet40  & 0.0005  & 0.2 & 64 &0 & 20 & 0.05 &16 \\\midrule\midrule
NTU2012*  & 0.01   & 0.2 & 256 & 1 & 0 & 0.05 & 16 \\
NTU2012*  & 0.01   & 0.2 & 256& 0 & 20 & 0.05 & 16 \\
\hline
ModelNet40*   & 0.001   & 0.2 & 512 &1 & 0 & 0.05& 16  \\
ModelNet40*  & 0.001   & 0.2 & 512 &0 & 20 & 0.05 & 16  \\
\hline
Yelp   & 0.01   & 0.2 & 64 & 1& 0 & 0.01 & 4   \\
Yelp   & 0.01   & 0.2 & 64 & 0& 1 & 0.01 & 4   \\
\hline
House(1)   & 0.01   & 0 & 512 &1 & 0 & 0.05 & 16   \\
House(1)  & 0.01   & 0.2 & 64 &0 & 1 & 1 & 16   \\
\hline
House(0.6) & 0.01   & 0 & 64&1 & 0 & 0.05 & 16 \\
House(0.6)   & 0.01   & 0.2 & 512&0 & 1 & 0.05 & 16 \\
\hline
Walmart(1)   & 0.01   & 0 & 256 & 80& 0 & 0.05 & 16 \\
Walmart(1)  & 0.001   & 0 & 256 & 0& 50 & 1 & 16 \\
\hline
Walmart(0.6)   & 0.01   & 0 & 256& 20 & 0 & 0.05 & 16  \\
Walmart(0.6)   & 0.001   & 0 & 256& 0 & 50 & 1 & 16  \\
\hline
20Newsgroups   & 0.01   & 0 & 64 &0.1 & 0 & 0.05 & 8   \\
20Newsgroups   & 0.01   & 0 & 64 &0 & 0.1 & 1 & 8   \\
\bottomrule\bottomrule
\end{tabular}\label{tab:hyperparams-phenomNN-H-ablation}
\end{table}

\end{document}